\documentclass{article}

\usepackage{PRIMEarxiv}

\usepackage[utf8]{inputenc} 
\usepackage[T1]{fontenc}    
\usepackage{hyperref}       
\usepackage{url}            
\usepackage{booktabs}       
\usepackage{amsfonts}       
\usepackage{nicefrac}       
\usepackage{microtype}      
\usepackage{lipsum}
\usepackage{fancyhdr}       
\usepackage{graphicx}       
\graphicspath{{media/}}     

\usepackage{multirow}
\usepackage{bm}
\usepackage{combelow}
\usepackage[numbers]{natbib}
\usepackage{float}
\usepackage{subfig} 
\usepackage{amsmath}
\usepackage{mathtools}
\usepackage{enumitem}
\usepackage[ruled,vlined,linesnumbered,noend]{algorithm2e}

\title{EDSA-Ensemble: an Event Detection Sentiment Analysis Ensemble Architecture}

\author{
  Alexandru Petrescu$^1$, Ciprian-Octavian Truic{\u{a}}$^1$, Elena-Simona Apostol$^1$, Adrian Paschke$^2$ \\
  $^1$University Politehnica of Bucharest, Bucharest, Romania \\
  $^2$Fraunhofer Institute for Open Communication Systems, Berlin, Germany \\
  \texttt{alex.petrescu@upb.ro, ciprian.truica@upb.ro, elena.apostol@upb.ro} \\
  \texttt{adrian.paschke@fokus.fraunhofer.de}
}

\begin{document}
\maketitle

\begin{abstract}
As global digitization continues to grow, technology becomes more affordable and easier to use, and social media platforms thrive, becoming the new means of spreading information and news. 
Communities are built around sharing and discussing current events.
Within these communities, users are enabled to share their opinions about each event.
Using Sentiment Analysis to understand the polarity of each message belonging to an event, as well as the entire event, can help to better understand the general and individual feelings of significant trends and the dynamics on online social networks.
In this context, we propose a new ensemble architecture, EDSA-Ensemble (Event Detection Sentiment Analysis Ensemble), that uses Event Detection and Sentiment Analysis to improve the detection of the polarity for current events from Social Media. 
For Event Detection, we use techniques based on Information Diffusion taking into account both the time span and the topics. 
To detect the polarity of each event, we preprocess the text and employ several Machine and Deep Learning models to create an ensemble model.
The preprocessing step includes several word representation models, i.e., raw frequency, $TFIDF$, Word2Vec, and Transformers.
The proposed EDSA-Ensemble architecture improves the event sentiment classification over the individual Machine and Deep Learning models.
\end{abstract}


\maketitle

\section{Introduction}\label{Introduction}

As social media platforms grow more and more each day, it also increases the need to analyze and understand certain aspects, such as the impact of important or spiking topics over the network\cite{petrescu2019sentiment}.
Event Detection techniques are used to automatically identify important or spiking topics by analysing social media data.
In this paper, we use the angle of the positive emotion generated by these topics for the users and the magnitude, both reach and time span, in order to better understand what is happening on social media platforms, mainly Twitter.

Sentiment Analysis is a field in Natural Language Processing that analyzes user opinions and emotions from written language~\cite{Medhat2014,Truica2021}, while Event Detection deals with analyzing information diffusion in graph networks~\cite{Guille2012}.
Although there is a large volume of work done on Event Detection using social media data and on Sentiment Analysis of this type of content, in the current literature, there is a shortcoming of the approaches that combine the two domains.
There are multiple communities that are involved in mining, gathering, and giving some meaning to the vast amount of content generated daily by the users of those platforms, namely the Network Analysis and Natural Language Processing communities.
The two communities are using different types of approaches since they have different purposes:
\begin{itemize}
 \item For the Network Analysis community, the main purpose is developing methods to deal with the spread and mitigation of harmful content using Event Detection. Event Detection is used to detect the impact and spread of topics on Social Networks using multiple types of approaches such as sliding windows, topic detection, etc. 
 \item For the Natural Language Processing community, the main purpose is to develop methods to analyze the opinion of the users using Sentiment Analysis. Sentiment Analysis uses supervised and unsupervised learning to determine the polarity of textual data. The dimensions of the actual classification of a text are different for each method, ranging from binary (positive/negative or neutral if we cannot tell) to multi-dimensional approaches (happiness, contempt, surprise, and so on). 
\end{itemize}

In this paper, we want to address this shortcoming and bridge the gap between the Network Analysis and Natural Language Processing communities by providing an Event Detection Sentiment Analysis Ensemble Architecture, i.e., EDSA-Ensemble.
EDSA-Ensemble first detects events from social media in almost real-time and then applies sentiment analysis on the resulting topic of each event using an ensemble approach.
By combining Network Analysis approaches with the detection of event sentiments using Machine and Deep Learning in Online Social Networks, our architecture brings together the research of these two isolated communities, i.e., Network Analysis and Natural Language Processing. 
We aim to evaluate individually various types of Event Detection strategies and Sentiment Analysis models and then combine them into an ensemble model to facilitate the detection and spread of events in a content-aware and network-aware manner.

The main research questions we want to address with this work are:
\begin{itemize}
    \item[$(Q_1)$] How can we determine the different events discussed on social media for a given time span?
    \item[$(Q_2)$] Can we accurately determine the individual opinion of users for a given event?
    \item[$(Q_3)$] Can we accurately determine the overall sentiment of these events?
\end{itemize}

To address these questions, we propose EDSA-Ensemble.
Firstly, EDSA-Ensemble uses multiple Event Detection algorithms to determine the important topics that are discussed online.
By using multiple algorithms that use different heuristics to detect topics of importance, we ensure that we do not miss any events for a given time span.
Secondly, EDSA-Ensemble uses multiple algorithms to determine the users' opinions by analyzing individual tweets.
Finally, EDSA-Ensemble determines the overall sentiment of events by aggregating through a voting process the results of the multiple individual sentiment analysis models, thus improving the accuracy of detecting the overall sentiment of an event.

The main objectives of our novel architecture are:
\begin{itemize}
    \item[$(O_1)$] Propose a modular design that determines events on social media using multiple algorithms.
    \item[$(O_2)$] Determine the individual user opinion regarding an event by developing multiple sentiment detection models.
    \item[$(O_3)$] Propose a voting-based ensemble model that determines with high accuracy the overall sentiment of an event.
\end{itemize}

Thus, the main contributions are as follows:
\begin{itemize}
    \item[$(C_1)$] We propose EDSA-Ensemble: a modular architecture that incorporates multiple Event Detection algorithms and Sentiment Analysis models to accurately detect the important events discussed on social media and their polarity.
    \item[$(C_2)$] We present an in-depth analysis of each module of the proposed architecture on a real-world dataset.
    \item[$(C_3)$] We do a thorough discussion on the obtained experimental results. 
\end{itemize}

This paper is structured as follows. 
Section \ref{RelatedWork} presents a survey of the current state-of-the-art methods for Event Detection and Sentiment Analysis.
Section \ref{Methodology} describes the EDSA-Ensemble architecture and its main functionalities.
Also, we briefly describe how each component interacts with the overall platform and how each chosen algorithm works.
Section \ref{ExperimentalResults} showcases the obtained results, analyzes them, and gives some directions for future trials using the chosen Event Detection and Sentiment Analysis methods.
Section \ref{Conclusions} concludes and presents several new directions and improvements for the proposed solution.

\section{Related Work}\label{RelatedWork}

In this section, we present an overview of the current state of the art related to Event Detection and Sentiment Analysis.

\subsection{Event Detection}

One issue of Event Detection tasks is the small amount of training data.
This leads to poor results for unseen/sparsely labeled trigger words, and the models are prone to overfitting densely labeled trigger words. 
To address this problem, \citet{Tong2020} propose a novel Enrichment Knowledge Distillation (EKD) model for leveraging external open-domain trigger knowledge to reduce the in-built biases to frequent trigger words in annotations. 
After running their experiments on ACE2005, they obtained better performance than the baseline, especially for the areas of interest, i.e., Unseen or Sparsely Labeled words.

When it comes to social data for event detection, current methods learn limited amounts of knowledge as they ignore the rich semantics and structural information of this type of dataset. 
And, sometimes, even they cannot memorize previously acquired knowledge. 
To address this issue, \citet{Cao2021} propose a novel Knowledge-Preserving Incremental Heterogeneous Graph Neural Network (KPGNN) for incremental social event detection by leveraging the power of GNNs.
KPGNN needs no feature engineering and requires a small amount of parameter tuning. 
They conducted experiments on MAVEN, a general domain event detection dataset constructed from Wikipedia documents, and obtained good results having better performance than the chosen baseline, which also contains LDA.

Another problem to tackle is collectively detecting multiple events, particularly in cross-sentence settings. 
For this, \citet{Lou2021} propose encoding the semantic information and using a model for event inter-dependency at a document level. 
They reformulate this problem as a Seq2Seq task and propose a Multi-Layer Bidirectional Network (MLBiNet) to capture the document-level association of events and semantic information simultaneously. 
Their solution obtains better performance on the ACE-2005 corpus than other state-of-the-art approaches by a few percentages in all tasks.

An additional issue for social media data for event detection is that the use of structural relationships among users is rarely observed in online Twitter network communities for event detection. 
To address this issue, \citet{Ansah2020} propose a new framework called SensorTree, which tracks information propagation in Twitter network communities to model spikes and later uses a tensorized topic model to extract the events. 
The experiments are conducted on 4 different Twitter datasets, and the proposed solution obtains better results than the baseline when using as metrics: Topic Intrusion Score, Topic Coherence, Precision, and ROGUE.

One more issue is that most frameworks used for event detection do not capture the embedded semantic meaning, which is usually highly heterogeneous and unstructured.
Also, most frameworks do not even identify relationships. 
To address this, \citet{Abebe2020} introduce a generic Social-based Event Detection, Description, and Linkage framework, i.e., SEDDaL. 
SEDDaL uses heterogeneous sources (e.g., Flickr, YouTube, and Twitter) for input and produces a collection of semantically meaningful events interconnected with spatial, temporal, and semantic relationships. 
SEDDaL outperformed qualitatively and quantitatively other popular methods on the MediaEvalSED 2013 and 2014 datasets.

Most previous research on automated event detection is focused only on statistical and syntactical features in data and is lacking the involvement of underlying semantics~\cite{Hettiarachchi2021}, which are important for effective information retrieval from text since they represent the connections between words and their meanings. 
\citet{Hettiarachchi2021} propose Embed2Detect, a method for event detection in social media that combines characteristics in word embeddings and hierarchical agglomerative clustering. 
They obtain better results than the other solutions, e.g., MABED.

Multi-source event propagation involves sharing and collaboration over the same topic from multiple people and there is not a lot of talk on this topic in the current literature. 
\citet{Shi2019} create a framework that uses previous knowledge about the event in order to obtain a multi-source events propagation model based on individual interest.
This solution is not compared with other approaches.

Traditional event detection methods mainly use sentence-level information to identify events and it usually uses topic modeling and topic labeling techniques~\cite{Truica2017,Truica2021b}.
However, the information used for detecting events is usually spread across multiple sentences, and sentence-level information is often insufficient to resolve ambiguities for some types of events. 
To address this challenge, \citet{Yao2021} propose FGCAN, a Filter-based Gated Contextual Attention Network model, which is augmented with hierarchical contextualized representations to utilize both sentence-level and document-level information. 
They ran experiments on the widely used ACE 2005 and KBP 2015 datasets and they manage to outperform the state-of-the-art approaches for this topic.

Due to the missing corpus labels or the limited performance of the classifier, sometimes sentiment time series may not correspond to the actual values, especially when the sentiment value changes drastically, called extreme value. 
To address this, \citet{Jingyi2021} propose a new method to calibrate the series for event detection based on the evaluation on a sampling dataset. 
The authors obtained good results compared with the proposed baseline.

EDGNN~\cite{Gao2021} is a new method that uses  Graph Neural Networks (GNN) for event detection.
The method starts with topic modeling, which is used to help the graph enrich the semantics of short texts, and continues with a text-level graph with fewer edges on top of BERT. 
A real-world foodborne disease event dataset is used to demonstrate that this model outperforms state-of-the-art baselines. 

Information diffusion is the field that analyses how the states of individual nodes in a graph evolve over time~\cite{Goswami2018}. 
It is used to detect user behavior and information spreading across social networks by monitoring the magnitude (e.g., the number of nodes it has reached) and the lifespan of a topic. 
In the case of Online Social Networks, these topics are named bursty topics~\cite{Diao2012}, i.e., topics that spread information better than the average, looking like a pike in the magnitude/time function representation for the topics. 

\citet{Guille2013} survey the state-of-the-art methods for Event Detection and ways of building a taxonomy with good performance (time spread and magnitude). 
Their analysis focuses on Twitter and detects bursty topics focusing on the evolution in time of the topics.
The analysis includes multiple algorithms and methods, e.g., 
Peaky Topics~\cite{Shamma2011} that uses normalized term frequency, 
TSTE (Temporal and Social Terms Evaluation)~\cite{cataldi2010emerging} that uses a five-step approach considering both temporal and social properties, 
MACD (Moving Average Convergence Divergence)~\cite{lu2012trend} that uses the trend momentum of a topic,
SDNML (Sequentially Discounting Normalized Maximum Likelihood)~\cite{takahashi2011discovering} specialized in tweets that also have media and URLs, and OLDA (Online Latent Dirichlet Allocation)~\cite{AlSumait2008} which is an improvement to the LDA adding, with the online phase.
Experimental results prove that the information is diffused by users that appear as the node for which the graph partitions~\cite{Guille2013Sondy}, the "influencers" can be equivalent to articulation points in the graph model.

\citet{Guille2012} propose an ML-based approach that uses inference of time-dependent diffusion probabilities from a multidimensional analysis of individual behaviors.
With this information, the solution builds communities (sub-graphs).
To have a good overview of information propagation, these sub-graphs are based on multiple dimensions: semantics, social, and time. 
\citet{Guille2014} present a new statistical method, i.e., Mention-Anomaly-Based Event Detection (MABED).
MABED relies on statistical measures computed from user mentions to detect events and their impact on Social Networks.
The approach builds a structure where, for each user, it takes at most two levels of followers to build communities, i.e., sub-graphs. 
As a result, it produces a list of events. 
Each event is described by one main word and a set of weighted related words, a period of time, and the magnitude of its impact on the crowd. 
The experimental results prove that MABED has higher topic readability and temporal precision than the algorithms used as the baseline, i.e., PS (Peakiness Score)~\cite{Shamma2011} and EDCoW (Event Detection with Clustering of Wavelet-based Signals)~\cite{Weng2011}.

At the other end of the spectrum, there are time-frame-based Event Detection methods such as Peaky Topics~\cite{Shamma2011}, which splits the corpus into equally distributed bins and detects the bursty topics in each bin. 
This method is particularly good for fast decision-making systems as it can stop the spread of harmful information in the networks considering the current window. The steps for this approach are as follows:
(1) split the data into equally distributed bins by time;
(2) in each bin, find the bursty topic using $TFIDF$; and
(3) combine the results, and in case of under the Threshold results, drop the bad topics and redo the steps with fewer bins.

\subsection{Sentiment Analysis} 
There are two main approaches for Sentiment Analysis: \textit{(1)} Lexicon Based (unsupervised methods that use word polarity to classify textual data), and \textit{(2)} Machine Learning (supervised methods that use polarity labeled dataset to build a model).

\citet{Sirbu2016} present a comprehensive study of the two main Sentiment Analysis approaches. 
The experiments show that the approach that achieves the best results uses Multivariate Analysis of Variance (MANOVA) to exact features. 
This method of feature extraction manages to highlight the relation between dependent and independent variables for the task of Sentiment Analysis.

\citet{Thelwall2011} try to determine the opinion presented in Twitter events during peak hours.
The results show that events with negative sentiments are more frequent, whereas positive predictions were accurate only during peak hours.
\citet{kouloumpis2011twitter} propose to use combine sentiment lexicons with the Twitter posts' linguistic features, i.e., n-grams, part-of-speech, etc., and metadata, i.e., hashtags, emoticons, and abbreviations for predicting tweets' polarity.
The experiments show that the models trained on the combination between the n-grams, lexicon, and metadata produce the best results.
\citet{pak2010twitter} observe that the overuse of adjectives influences the accuracy of sentence-level sentiment detection.
By creating a Naïve Bayes model that takes into account the polarity of adjectives, they manage to improve the overall performance of the sentence-level sentiment detection task.

In the current literature, deep learning models that employ Convolutional Neural Networks (CNN), Recurrent Neural Networks (RNN), Gated Recurrent Units (GRU), and Long Short-Term Memory (LSTM) for text classify ~\cite{kalchbrenner2014convolutional} are also used for sentiment detection.
\cite{Phan2020} study how a feature ensemble model related to tweets containing fuzzy sentiment can improve polarity detection.
By considering the lexical, word-type, semantic, position, and sentiment polarity of words, the authors create a Tweet Embedding used as the input for a CNN that improves the performance of tweet sentiment analysis in terms of the F1 score.
\citet{Li2022} propose BiERU (Bidirectional emotional recurrent unit for conversational sentiment analysis) that is able to extract the context from conversations and improve sentiment analysis.
BiERU is a fast, compact, and parameter-efficient party-ignorant framework.
\citet{Yang2020} introduce the SLCABG model, a deep learning model that employs sentiment lexicons and a deep learning architecture with CNN and attention-based Bidirectional GRU (BiGRU) to determine the sentiments from reviews.
\citet{Onan2020} proposes a CNN-LSTM architecture with a TFIDF weighted GloVe~\cite{Pennington2014} embedding, while \citet{Ruz2020} use a Bayesian network classifiers for sentiment analysis.
\citet{Behera2021} also combines CNN and LSTM to extract the opinion from consumer reviews posted on social media.
\citet{Naseem2020} propose DICE, a transformer-based method for sentiment analysis.
DICE encodes representation from a transformer and applies deep intelligent contextual embedding to enhance the quality of tweets by removing noise.
The model uses the word sentiments, polysemy, syntax, and semantic knowledge together with a BiLSTM classifier to extract tweets' polarity.

Another research direction is aspect-based sentiment analysis (ABSA) which tries to perform a multi-faceted analysis of textual data in order to determine the sentiment regarding both target items as well as target categories.
\citet{Basiri2021} propose an Attention-based Bidirectional CNN-RNN Deep Model (ABCDM).
The experiment results show that ABCDM produces state-of-the-art performance on eight datasets, five on reviews and three on Twitter data.
\citet{Peng2020} propose an aspect sentiment triplet extraction deep neural model for the Aspect-Based Sentiment Analysis task. 
Their technique uses triplet extraction, i.e., (What, How, Why) from the inputs, which show \textit{What} the targeted aspects are, \textit{How} their sentiment polarities are, and \textit{Why} they have such polarities.

\subsection{Event Sentiment Analysis}

In the current literature, event sentiment analysis has not yet been widely explored, although it is a very important topic~\cite{Ebrahimi2017}.

\citet{Fukuhara2007} propose a new method that creates two types of graphs to show temporal changes in topics and sentiments.
To visualize temporal changes in topics associated with sentiments, topic graphs are created.
Sentiment graphs are used to show the temporal change of sentiments associated with a topic.
The experiments use news articles to visualize these types of graphs.

\citet{Zhou2013} propose a Tweets Sentiment Analysis Model (TSAM) that extracts the societal interest and users' opinions regarding a social event.
TSAM uses a lexicon to extract the individual polarity and, after preprocessing the text and identifying the entities, an overall sentiment score is calculated.
The experiments focus on extracting the users' opinions regarding specific political candidates.

\citet{Makrehchi2013} propose a new social media text labeling approach that considers stock market events.
The authors extract and collect stock movement information from Twitter and assign a polarity label for each tweet.
They train a classification model using individual tweets to predict the polarity and then aggregate the results to obtain a net sentiment for each day.
The proposed model obtains a 0.96 F1-Score.

\citet{Patil2018} propose SegAnalysis, a framework that performs tweet segmentation, event detection, and sentiment analysis.
Naïve Bayes and online clustering are used to detect events.
An overall sentiment score is computed for each event using a lexicon approach.
SegAnalysis remains a theoretical framework as no experiments are performed.

\citet{petrescu2019sentiment} propose the use of Logistic Regression and Support Vector Machines to analyze individual events extracted by MABED and OLDA.
The experimental results prove that the proposed model has a high accuracy in detecting the overall event sentiment.

\cite{Basu2022} introduce a new feature set selection framework. 
The feature set uses Information Gain to extract topics from live-streamed tweets.
Bag-of-words with N-Gram identification is applied to extract event features from the textual data as well as part-of-speech linguistic annotation.
A decision tree is used to extract the tweets that belong to a topic from the live Twitter stream and their polarity.
The experiments show that the proposed approach obtains a training accuracy of 92.62\%.

\citet{Zhang2022} propose an End-to-End Event-level Sentiment Analysis (E3SA) approach to solve the task of structured event-level sentiment analysis.
To enhance the event-level polarity detection, the method explicitly extracts and models the event structure information, i.e., subject, object, time, and location.
For the experiments, the authors collect and label their own dataset in Chinese texts from the finance domain.
E3SA obtains an accuracy of 86.17\% for the event-level sentiment analysis task on the collected dataset.

\section{Methodology}\label{Methodology}

In this section, we present our proposed solution for accurate event and sentiment detection in social media. 
This solution offers two classes of tasks: Event Detection and Sentiment Analysis tasks. 
For the Event Detection task, we are using a raw frequency-based embedding as a method to vectorize the text. 
For the Sentiment Analysis task, we are using the standard architectures for the chosen methods.

\subsection{Proposed Architecture}

\begin{figure}[H]
 \centering
 \includegraphics[width=1\columnwidth]{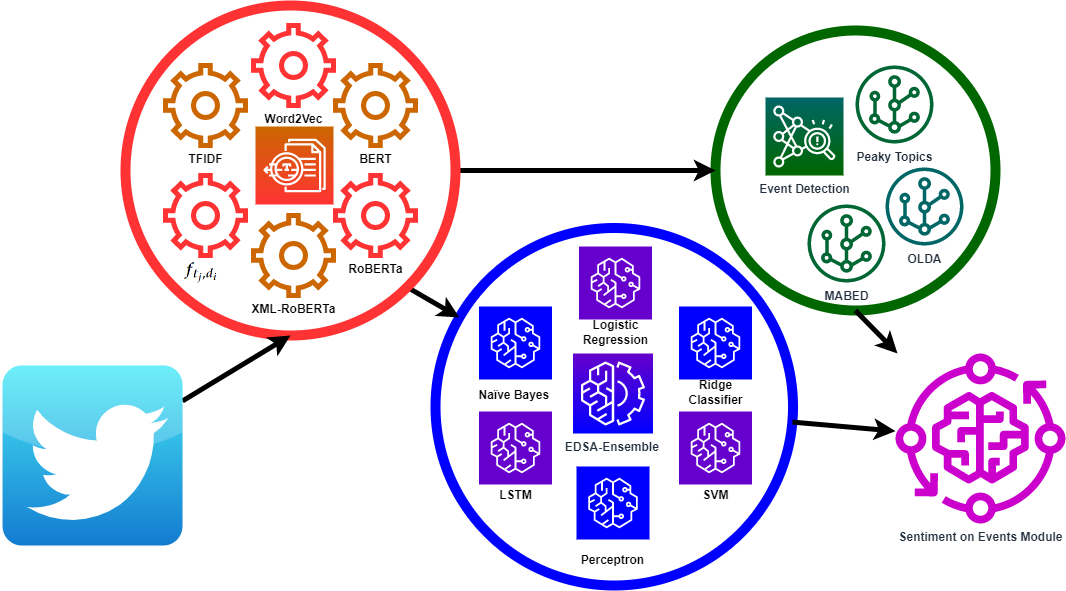}
 \caption{EDSA-Ensemble architecture diagram}
 \label{fig:ModuleWF}
\end{figure}

The architecture of the proposed solution EDSA-Ensemble is presented in Figure~\ref{fig:ModuleWF}.
Its main functionality and modules are as follows:
\begin{itemize}
 \item \emph{Twitter Collection Module} - is constantly listening for Twitter data in order to fetch and pass it on to the Preprocessing Module.
 \item \emph{Preprocessing Module} - applies different models for text representation on the received datasets from the Collection Module.
 \item \emph{Event Detection Module} - is used to detect bursty topics either on cleaned or raw text.
 \item \emph{Sentiment Analysis Module} - applies different methods for sentiment analysis on the preprocessed datasets, e.g., Classical ML Models (Naïve Bayes, Support Vector Machines) and Deep Models (BiLSTM and Transformers), and uses an ensemble voting system to determine the correct class for each document.
  \item \emph{Sentiments on Events Module} - applies Sentiment Analysis methods on the outputs of the Event Detection Module in order to determine if an event has a positive or negative impact over the network.
\end{itemize}

In the following subsections, we present a detailed description of each architectural module from Figure~\ref{fig:ModuleWF}.

\subsection{Preprocessing Module}

In Natural Language Processing, words are represented using numerical values or weight. 
Thus, a weight (numerical value or vector) is assigned to a word. 
These weights can encapsulate the importance of a word in a corpus of text and also its context. 
We employ three models for representing terms from the corpus 
\textit{(1)} Bag-of-Words model using raw frequency $f_{t_{j},d_{i}}$ and $TFIDF$;
\textit{(2)} Word Embedding model using Word2Vec; and
\textit{(3)} Transformer Embedding model using BERT, RoBERTa, and XLM-RoBERTa.

\subsubsection{Bag-of-Words Model}

The Bag-of-Words model can be built using multiple weighting schemes.
Given a corpus of documents $D = \{ d_{i} \mid i \in \overline{1,n} \}$ ($n=\mid\mid D \mid\mid$ is the size of the documents) and a vocabulary $V = \{ t_{j} \mid j \in \overline{1,m} \}$ ($m=\mid\mid V \mid\mid$ is the size of the vocabulary) that contains the set of unique terms $t_{j}$ that appear in dataset, we can define using the Bag-of-Words model multiple weighting schemes that encode documents and terms into vectors.
EDSA employ two of them: 
\textit{i)} the raw frequency and ($TF$);
\textit{ii)} term frequency-inverse document frequency ($TFIDF$).

The raw frequency ($f_{t_{j},d_{i}}$) counts the number of co-occurrences of a word in a document.
It provides no context and adds bias toward longer documents.
The term frequency-inverse document frequency ($TFIDF$) weighting scheme is often used as a central tool for scoring and ranking terms in a document.
The $TFIDF$ weight (Equation~\eqref{eq:tfidf}) is computed by multiplying the augmented term frequency ($TF$) with the inverse document frequency ($IDF$).
For $TFIDF$, the importance increases proportionally to the number of times a term appears in the document but is offset by the frequency of the term in the corpus.
To prevent a bias towards long documents, $TF$ (Equation~\eqref{eq:tf}) is computed as the number of co-occurrences $f_{t_{j},d_{i}}$ of a word in a document normalized by the length of the document ($\sum _{t'\in d}{f_{t',d}}$) when computing the $TF$~\cite{Paltoglou2010}. 
$IDF$ (Equation~\eqref{eq:idf}) is used for minimizing the importance of common terms that bring no information value by reducing the TF weight by a factor that grows with the collection frequency $n_{j}$ of a term $t_{j}$.

\begin{equation}
\label{eq:tfidf}
TFIDF(t_{j}, d_{i}, D) = TF(t_{j}, d_{i}) \cdot IDF(t_{j}, D)
\end{equation}

\begin{equation}
\label{eq:tf}
TF(t_{j}, d_{i}) = \frac{f_{t_{i},d}}{\sum _{t'\in d}{f_{t',d_{i}}}}
\end{equation}

\begin{equation}
\label{eq:idf}
IDF(t_{j}, D) = \log {\frac {n}{n_{j}}}
\end{equation}

Using the term weights, we can construct a document-term matrix $A = \{ w_{ij} \mid i = \overline{1,n} \wedge j = \overline{1,m} \}$, where rows correspond to documents and terms to columns.
The cell value $w_{ij}$ is the weight (e.g., $f_{t_{i},d}$, $\textsc{TF}$, $\textsc{TFIDF}$, etc.) of term $t_{j}$ in document $d_{i}$.

\subsubsection{Word Embedding Model}

Word embedding consists of a set of techniques used for weighting word and text analysis.
It features learning techniques that assign numerical values (usually as a vector) to linguistic words.
The proposed solution uses the Word2Vec technique to construct a Common Bag Of Words (CBOW) model~\cite{mikolov2013efficient}, which is a type of word embedding.

The CBOW model tries to predict the target word by looking at the words situated at its left and right sides. The model uses one-hot encoder vectors of size $m$ as input context $(x_{-m}, ..., x_{-1}, x_{1},..., {x_m})$ for each target word $x_{i} \in {\!R}^{m}$. The embedding word vectors for the context are $v_{i} = Vx_{i}$, where $V$ is the input word matrix and $i \in \{-m,..., -1, 1, ..., m\}$.

The dot product of vectors is used to push similar words close to each other and maximize the model's performance. The output score is computed using the average of all word embedding vectors as $Z=U\hat{v}$, where $U$ is the output word matrix and $\hat{v}$ is the average of $m$ context vectors for the target word target~\cite{Singh2018}. The output is a word vector ($\hat{y}$) obtained from the softmax function (Equation~\eqref{eq:W2VCBOW}).

\begin{equation}
\label{eq:W2VCBOW}
\hat{y} = \frac{e^{\hat{y}}}{\sum_{k=1}^{\mid V \mid} e^{\hat{y}_k}}
\end{equation}

\subsubsection{Transformers}

In machine learning, a transformer is a deep learning model that adopts the mechanism of self-attention which balances the importance of each token in the input data. Transformers were introduced in 2017 by a team at Google Brain and they became a popular approach in NLP tasks, replacing RNN models such as long short-term memory (LSTM), obtaining state-of-the-art performance with the rise of BERT\cite{Devlin2019}. Other popular approaches such as RoBERTa~\cite{Liu2019} and XLM-RoBERTa\cite{Conneau2020} have the same architecture but have different training data which result in a different performance for each task that they are not specialized in.

We are going to benchmark the previously mentioned transformers against our Twitter data since transformers obtained state-of-the-art performance in NLP tasks and we are using multiple variations in order to see how the increasing quantity of data, RoBERTa, affects the base, BERT, or how the multilingual features, XLM-RoBERTa, do.

\paragraph{BERT}

BERT (Bidirectional Encoder Representations from Transformers)\cite{Devlin2019} proves the importance of bidirectional pre-training by surpassing the previous state-of-the-art approach which was a left-to-right approach, while BERT instead uses the self-attention mechanism to unify these two stages, as encoding a concatenated text pair with self-attention effectively includes bidirectional cross attention between two sentences. For this paper, pre-trained embeddings used are the WordPiece embeddings that contain a vocabulary of 30,000 tokens and a pre-training on the following datasets 
\textit{(1)} BooksCorpus (800M words) and
\textit{(2)} English Wikipedia (2,500M words) using everything except lists, tables, and headers.

The core BERT framework has 2 major steps:
\begin{enumerate}
    \item Pre-training BERT:  Masked LM bidirectional training with randomly masked tokens (15\% of the input) trying to predict those tokens, the only downside being that the mask is not present during the fine-tuning. In order to mitigate this for the randomly chosen tokens 80\% of the time the swap is done with the mask, 10\% with another random token and 10\% of the time it is left as it is. Next Sentence Prediction is used for Q\&A modelling where 50\% of the time the sequence is correct.
    \item Fine-tuning: A straightforward strategy since the self-attention mechanism in the Transformer allows BERT to model many downstream tasks.
\end{enumerate}

\citet{Devlin2019} state that it is critical to use a document-level corpus rather than a shuffled sentence-level corpus such as the Billion Word Benchmark~\cite{chelba2014one} in order to extract long contiguous sequences.

\paragraph{RoBERTa}

RoBERTa (Robustly Optimized BERT Pretraining Approach)~\cite{Liu2019} is a retraining of BERT with improved training methodology, more data, and more computing power.
To improve the training procedure, RoBERTa removes the Next Sentence Prediction (NSP) task from BERT’s pre-training and introduces dynamic masking so that the masked token changes during the training epochs. Larger batch-training sizes were also found to be more useful in the training procedure.
Importantly, RoBERTa uses 160 GB of text for pre-training, including 16GB of Books Corpus and English Wikipedia used in BERT. 
The additional data included CommonCrawl News dataset (63 million articles, 76 GB), Web text corpus (38 GB), and Stories from Common Crawl (31 GB). 
This coupled with a whopping 1024 V100 Tesla GPUs running for a day, led to pre-training of RoBERTa.

\paragraph{XLM-RoBERTa}

XLM-RoBERTa~\cite{Conneau2020} model is a transformer-based pre-trained multilingual masked language model able to determine the correct language just from the input ids, being pre-trained on 2.5TB of filtered CommonCrawl data containing 100 languages, mainly being a multilingual version or RoBERTa.

Due to the fact that we are using social data for this task, we might encounter multilingual content even if the selected language was English and for this particular reason, another transformer seems like a good benchmark.

\subsection{Event Detection Module}

For Event Detection, we use the Mention-Anomaly-Based Event Detection (MABED)~\cite{Guille2014}, Online LDA (OLDA)~\cite{AlSumait2008}, and Peaky Topics~\cite{Shamma2011} algorithms.
All approaches detect bursty topics either on cleaned or raw text. The cleaned text results by removing the stop words and any non-character from the raw text and applying lemmatization.

\subsubsection{Mention-Anomaly-Based Event Detection}

\textit{Mention-Anomaly-Based Event Detection (MABED)} is an efficient statistical method used for detecting events in social networks since it is immune to social media bias consisting of unrelated texts on a given topic~\cite{Guille2014}.
Although social networks can contain spam messages with no actual intent posted around certain hours, this method is efficient when filtering irrelevant content.
MABED is not always viable since, without external context, it can sometimes distort some types of events. 

To identify a bursty topic and detect an event for a period of time $I = [ a ; b]$ and a main word $t$, a weight $w_{t'_{q}}$ is computed for each candidate word $t'_{q}$ in the time slice $i$ (Equation~\eqref{eq:mabedb}).
The affine function $\rho_{O_{t, t'_{q}}}$ (Equation~\eqref{eq:mabed_rho}) is used to compute the weight.
This function corresponds to the first order auto-correlation of the time series for $N_{t}^{i}$ (number of tweets in the time-slice $i$ that contain the main word $t$) and $N_{t'_{q}}^{i}$ (number of tweets in the time-slice $i$ that contain the candidate word $t'_{q}$)~\cite{Erdem2014}.

\begin{equation}\label{eq:mabedb}
 w_{t'_{q}} = \frac{\rho_{O_{t, t'_q}} + 1}{2}
\end{equation}

\begin{equation}\label{eq:mabed_rho}
 \rho_{O_{t, t'_{q}}} = \frac{\sum_{i = a + 1}^{b} A_{t, t'_{q}}}{(b - a - 1) A_{t} A_{t'_{q}}}
\end{equation}

Where:
\begin{enumerate}
 \item $A_{t,t'_{q}} = (N_{t}^{i} - N_{t}^{i-1})(N_{t'_{q}}^{i} - N_{t'_{q}}^{i-1})$
 \item $A_{t}^{2} = \frac{\sum_{i = a + 1}^{b} (N_{t}^{i} - N_{t}^{i - 1})^2}{(b - a - 1)}$;
 \item $A_{t'_q}^2 = \frac{\sum_{i = a + 1}^{b} (N_{t'_q}^{i} - N_{t'_q}^{i - 1})^2}{(b - a - 1)}$;
\end{enumerate}

\subsubsection{Online LDA}

Online LDA~\cite{AlSumait2008}, is an update on Latent Dirichlet Allocation~\cite{Blei2003} which uses a non-Markov on-line LDA Gibbs sampler topic model.
This model is capable of detecting bursty topics by capturing thematic patterns and identifying emerging topics and their changes over time.
This approach is sometimes better than the original LDA approach, as it updates the previous model at each time frame.

OLDA is a hierarchical Bayesian network that relates words and documents through latent topics using a probabilistic formula for a given term to be assigned to a given topic in the (current) context Equation~\eqref{eq:olda}.

\begin{equation}\label{eq:olda}
P(z_i = j \mid z_{\neg_i} ,w_{d_i}, \alpha, \beta) \propto \frac{C_{w_{\neg_i},j}^{V K} + \beta_{w_i,j}}{\sum_{v=1}^{V}{C_{v_{\neg_i},j}^{V K} + \beta_{v,j}}} \times \frac{C_{d_{\neg_i},j}^{D K} + \alpha_{w_i,j}}{\sum_{v=1}^{V}{C_{d_{\neg_i},j}^{D K} + \alpha_{d,j}}} 
\end{equation}

Where:
\begin{itemize}
 \item $D$ is the total number of documents; 
 \item $K$ is the number of topics;
 \item $V$ is the total number of unique words;
 \item $C_{w_{\neg_i},j}^{V K}$ is the number of times word $w$ is assigned to topic $j$, not including the current token instance $i$
 \item $C_{d_{\neg_i},j}^{D K}$ is the number of times topic $j$ is assigned to some word token in document $d$, not including the current instance $i$;
 \item $z_{\neg i}$ are all other word tokens; 
 \item $w_{d_i}$ are the unique word associated with the $i$-th token in document $d$ at the current time;
 \item $\beta_k$ is the $V$ vector of priors for topic $k$ at the current time;
 \item $\alpha_k$ is the $K$ vector of priors for document $k$ at the current time.
\end{itemize}

\subsubsection{Peaky Topics}

Identifying peaks in a time series is a very important task in many cases when immediate action is required which may point out \textit{1)} high demands in power distribution data, \textit{2)} increased CPU utilization, \textit{3)} burst in traffic or \textit{4)} hasty price increasing in financial data. 

For the particular case of stream-series on Twitter, let us consider $S = \{t_1, t_2, ..., t_M\}$ a text steam collected during a time period.
To apply the Peaky Topics algorithm, this stream should be divided in multiple equally distributed time-frame based bins~\cite{Shamma2011}.
After the splitting, we obtain $T = \{x_1, x_2, ..., x_N\}$ where $x_i \in T$ represents a tweet posted in the time-frame $T$.
The size of the bin, time-window, should be adapted relative to the total duration of the initial text stream.

For every $x_i$ value in the $T$, a check value is calculated and is used to determine if a point represents a local maximum or not.
To determine the spikes in a given frame, we consider many small sub-bins.
For each sub-bin we run a spike detection algorithm based on the number of tweets in those sub-bins.

Peaky Topics performs better at detecting bursty topics in narrow time frames compared to MABED and OLDA, but it does not take into consideration the evolution in time of the topic~\cite{Diao2012}.

\subsection{Sentiment Analysis Module}

For Sentiment Analysis, we use Naïve Bayes (NB), Logistic Regression (LR), Ridge Classifier (RC), Support Vector Machines (SVM), and Long Short-Term Memory Neural Networks (LSTM).
All approaches behave well on both raw texts or with feature engineering, and negation fusion.
Each method has an advantage over the other either in run-time, particular benchmarks, or overall benchmarks.

\subsubsection{Naïve Bayes}

Naïve Bayes (NB) is one of the base methods for sentiment analysis with good results most of the time, in our paper is the baseline.
NB is a probabilistic classification algorithm that computes the probability $ p(y = y_{\kappa} \mid d = d_{i})=p(y = y_{\kappa} \mid t_{1} = t_{i1}, ..., t_{m} = t_{im})$ of a document $d_{i} = \{ t_{ij} \mid j = \overline{1, m} \}$ to a given class $y_{\kappa} \in C$ ($i = \overline{1, n}$) using the Bayes Theorem.
Equation~\eqref{eq:nb} presents the Bayes Theorem, where $p(d = d_{i})$ and $p(y = y_{\kappa})$ are the probability of a document, respectively a class, and $p(d = d_{i} \mid y = y_{\kappa})=p(t_{1}=t_{i1}, ..., t_{m}=t_{im} \mid y = y_{\kappa})$ is the probability of class $y_{\kappa}$ given a document $d_{i}$.

\begin{equation}
\label{eq:nb}
 p(y = y_{\kappa} \mid t_{1} = t_{i1}, ..., t_{m} = t_{im}) = \frac{p(y = y_{\kappa})p(t_{1} = t_{i1}, ..., t_{m} = t_{im} \mid y = y_{\kappa})}{p(d = d_{i})}
\end{equation}

The denominator $p(d = d_{i})$ is constant and the numerator is equivalent to the joint probability $p(y = y_{\kappa}, t_{1} = t_{i1}, ..., t_{m} = t_{im})$.

Furthermore, all the terms are conditionally independent given a class $y_{\kappa}$, thus $p(y = y_{\kappa}, t_{1} = t_{i1}, ..., t_{m} = t_{im}) = p(y =y_{\kappa})\prod^{j=1}_{m}p(t_{j} = t_{ij} \mid y = y_{\kappa})$ and the model $p(y = y_{\kappa} \mid t_{1} = t_{i1}, ..., t_{m} = t_{im}) \propto p(y = y_{\kappa}, t_{1} = t_{i1}, ..., t_{m} = t_{im})$.

The Naïve Bayes classifier tries to estimate the class $\hat{y}_{i}$ for a document $d_{i}$ using Equation~\eqref{eq:nb_estimator}. 

\begin{equation}\label{eq:nb_estimator}
 \hat{y}_{i} = {argmax}_{y_{\kappa} \in C} p(y = y_{\kappa})\prod^{j=1}_{m}p(t_{j} = t_{ij} \mid y = y_{\kappa})
\end{equation}

\textit{Multinomial Naïve Bayes} uses a multinomial representation to model the distribution of words in a document. 
Thus, this model uses the distribution of probabilities to determine the class where each word appears frequently (Equation~\eqref{eq:mnb_pwc}).
The model makes two assumptions~\cite{Rennie2003}: 1) a document is a sequence of words and 2) the position of a word is generated independently.
The model uses the co-occurrence $f_{t_{j}, d_{i}}$ of a term $t_{j}$ in the document $d_{i}$.
Equation~\eqref{eq:mnb} presents the Multinomial Naïve Bayes model, while Equation~\eqref{eq:mnb_yest} presents the estimated values $\hat{y}_i$.

\begin{equation}\label{eq:mnb_pwc}
 p(t_{1} = t_{i1}, ..., t_{m} = t_{im} \mid y = y_{\kappa}) = \frac{(\sum_{j=1}^{m}f_{t_{j}, d_{i}})!}{\prod_{j=1}^{m}f_{t_{j}, d_{i}}!}\prod_{j=1}^{m}p(t_{j} = t_{ij} \mid y = y_{\kappa})^{f_{t_{j}, d_{i}}}
\end{equation}

\begin{equation}\label{eq:mnb}
p(y = y_{\kappa} \mid t_{1} = t_{i1}, ..., t_{m} = t_{im}) \propto 
p(y = y_{\kappa}) \frac{(\sum_{j=1}^{m}f_{t_{j}, d_{i}})!}{\prod_{j=1}^{m}f_{t_{j}, d_{i}}!}\prod_{j=1}^{m}p(t_{j} = t_{ij} \mid y = y_{\kappa})^{f_{t_{j}, d_{i}}}
\end{equation}

\begin{equation}\label{eq:mnb_yest}
 \hat{y}_{i} = {argmax}_{y_{\kappa} \in C} p(y_{i} = y_{\kappa}) \frac{(\sum_{j=1}^{m}f_{t_{j}, d_{i}})!}{\prod_{j=1}^{m}f_{t_{j}, d_{i}}!}\prod_{j=1}^{m}p(t_{j} = t_{ij} \mid y = y_{\kappa})^{f_{t_{j}, d_{i}}}
\end{equation}

\subsubsection{Logistic regression}

Logistic regression (LR)~\cite{hosmer2013} is a statistical model, widely used, that in its basic form uses a logistic function to model a binary classification for the given dataset. It tries to build a function, a linear one for this case that summarizes the input data. 
Logistic regression is a supervised method that behaves better than Linear Regression~\cite{montgomery2012introduction} when applied to a sparse dataset,
and builds classification models within a probabilistic context~\cite{Adeli2019}.
We consider that a dataset is sparse when the matrix of its representation has a lot of zeros and a few one values.
If we represent the words depending on their appearance into a certain block of text (tweet) or not,
we obtain a sparse matrix since the tweets have at most $140$ characters.

Given a set of classes $C=\{y_{\kappa} \mid \kappa = \overline{1, k} \}$ and a document dataset $D = \{d_{i} \mid i = \overline{1, n} \}$ described by the document-term matrix $A$, the model computes the probability of a class $y_{\kappa}$ given a document $d_{i}$ as $p(y = y_{\kappa} \mid d = d_{i})$.
The probability is the sigmoid function that maps documents to classes in order to determine the parameters of the vector $\bm{\beta} = \{ \beta_{0}, \beta_{1}, \cdots, \beta_{m} \}$ that fit the regression line.
Equation~\eqref{eq:lr} presents the Logistic Regression probability, where $x_{i} = [ 1 ] \oplus a_{i}$ is a vector composed by concatenating (operator $\oplus$) a one element vector with the value $1$ to the line $a_{i}$ to accommodate the $\beta_{0}$.
The vector $a_{i}$ is the corresponding line in matrix $A$ to document $d_{i}$. 
The one-element vector is used to accommodate the intercept of the regression line, i.e., $\beta_{0}$.

\begin{equation}\label{eq:lr}
 p(y = y_{\kappa} \mid d = d_{i}) = \frac{1}{1 + e^{-\bm{\beta}^{T}x_{i}}}
\end{equation}

To estimate the classes $\hat{y}_{i} = {argmax}_{y_{\kappa} \in C} p(y = y_{\kappa} \mid d = d_{i})$ and to determine the parameters $\bm{\beta}$ that give the best results, the algorithm maximises the log-likelihood using gradient descent~\cite{Yi2019}. 
The log-likelihood function guarantees that the gradient descent algorithm can converge to the global minimum.
Equation~\eqref{eq:loglike} presents the log-likelihood function $l(\bm{\beta})$, where $y_{i}$ is the real class of document $d_{i}$.
 
\begin{equation}\label{eq:loglike}
 l(\bm{\beta}) = \sum_{i=1}^{n} \left( y_{i} x_{i} \bm{\beta} - \log(1+e^{x_{i}\bm{\beta}}) \right)
\end{equation}

\subsubsection{Ridge classifier}

The Ridge Classifier (RC)~\cite{Pereira2016,Singh2016} or the Ridge regression is a learning approach that first normalizes the target values in the range $\{-1, 1\}$ and then treats the problem as a regression task (multi-output regression in the multi-class case).
Experimentally this method gives results at least as good as the Logistic Regression.
The cost function of the Ridge Regression is adjusted by adding a penalty $\lambda$ comparable to the square of the magnitude of the coefficients.
Equation~\eqref{eq:ridge-classifier} presents the log-likelihood for the Ridge Regression.

\begin{equation}\label{eq:ridge-classifier}
 l(\bm{\beta}) = \sum_{i=1}^{n} \left( y_{i} x_{i} \bm{\beta} - \log(1+e^{x_{i}\bm{\beta}}) \right) - \lambda\sum_{j=0}^{m} \beta_{j}^{2}
\end{equation}

\subsubsection{Support vector machines}

Support vector machines (SVM)~\cite{cortes1995support} is a supervised learning algorithm used in ML for classifications. 
Given a tagged dataset, SVM builds a function that separates the space of this given dataset into two classes.
In Sentiment Analysis, this type of model is used for binary classification, where the classes are positive and negative representing the tweets' sentiments. 
For the Sentiment Analysis tasks, we employ a linear SVM.
The SVM estimates the class $\hat{y}_{i} = {argmax}_{y_{\kappa} \in C}f(x_{i}, y_{\kappa})$.
Equation~\eqref{eq:SVMTrain} is used to separate the values of the training set using into the positive (+1) and negative classes (-1), where $w^{T}_{i}$ is the weight vector and $b$ is the bias.

\begin{equation}\label{eq:SVMTrain}
 f(x_{i}, y) = sign(w^{T}_{i} x_{i} + b)
\end{equation}

\subsubsection{Perceptron}

A perceptron is a type of linear classifier used in neural networks usually in a final layer such as softmax which uses the normalized exponential function, equation \ref{eq:softmax}, as an activation function to distinguish between the classes of the binary classification. The standard representation is given by Equation~\ref{eq:perceptron}, where $x_t \in {\rm I\!R}^m$ is the input of the perceptron with each dimension being a feature of the given input and $y_t = f(x_t)$ being the class, where $\bm{w}$ is a vector of real-valued weights, $\bm{w} \cdot x_{t}$ is the dot product $\sum _{i=1}^{m}w_{i}x_{ti}$, $b$ is the bias, and $\sigma(z)$ is the activation function.

\begin{equation}\label{eq:perceptron}
    \hat{y}_t = f(x_{t}) = \sigma(\bm{w} x_{t} + b)
\end{equation}

The input of the perceptron is a vector that is weighted in the activation function in order to produce an output in the \([0,1]\) domain usually being combined with a threshold to give the predicted class $\hat{y}_t$ for the input $x_{t}$.
In our architectures, we are using the softmax activation function (Equation~\ref{eq:softmax}) as the final layer for predicting the class $\hat{y}_t$.

\begin{equation}\label{eq:softmax}
    \sigma(x_t) = \frac{e^{\bm{w} x_t}}{\sum_{j=1}^m e^{\bm{w} x_{tj}}}
\end{equation}

\subsubsection{Long short-term memory}

A Neural Network based approach for text classification with good results is Long short-term memory (LSTM)~\cite{sachan2019revisiting}.
LSTM is an artificial recurrent neural network (RNN) architecture used in the field of deep learning. Unlike standard feed-forward neural networks, an LSTM has feedback connections. 
It can not only process single data points so it has been adapted to work for sentiment analysis. 

The LSTM uses two state components for learning the current input vector: \textit{(1)} a short-term memory component represented by the hidden state, and \textit{(2)} a long-term memory component represented by the internal cell state. 
The cell also incorporates a gating mechanism with three gates: \textit{(1)} input gate ($i_t \in \mathbb{R}^{n}$), \textit{(2)} forget gate ($f_t \in \mathbb{R}^{n}$), and \textit{(3)} output gate ($o_g \in \mathbb{R}^{n}$). 
The cell state corresponds to the long-term memory. 
LSTM avoids the vanishing and the exploding gradient issues. Equation~\eqref{eq:lstm} presents the compact forms for the state updates of the LSTM unit for a time step $t$, where:
\begin{itemize}
 \item $x_t \in \mathbb{R}^m$ is the input features vector of dimension $m$ ; 
 \item $h_t \in \mathbb{R}^n$ is the hidden state vector as well as the unit's output vector of dimension $n$, where the initial value is $h_0=0$;
 \item $\Tilde{c}_t \in \mathbb{R}^n$ is the input activation vector;
 \item $c_t \in \mathbb{R}^n$ is the cell state vector, with the initial value $c_0=0$;
 \item $W_i, W_f, W_o, W_c \in \mathbb{R}^{n \times m}$ are the weight matrices corresponding to the current input of the input gate, output gate, forget gate, and the cell state;
 \item $V_i, V_f, V_o, V_c \in \mathbb{R}^{n \times n}$ are the weight matrices corresponding to the hidden output of the previous state for the current input of the input gate, output gate, forget gate, and the cell state;
 \item $b_i, b_f, b_o, b_c \in \mathbb{R}^{n}$ are the bias vectors corresponding to the current input of the input gate, output gate, forget gate, and the cell state;
 \item $\delta_s(z)=\sigma_{g}(z)=\frac{1}{1+e^{-z}} \in [0, 1]$ is the sigmoid activation function;
 \item $\delta_h(z)=\tanh(z)=\frac{e^{z} - e^{-z}}{e^{z} + e^{-z}} \in [-1, 1]$ is the hyperbolic tangent activation function;
 \item $\odot$ is the element-wise product, i.e., Hadamard Product.
\end{itemize}

\begin{equation}\label{eq:lstm}
\begin{split}
i_t & = \delta_{s}(W_i x_{t} + V_i h_{t-1} + b_i) \\
f_t & = \delta_{s}(W_f x_{t} + V_f h_{t-1} + b_f)\\
o_t & = \delta_{s}(W_o x_{t} + V_o h_{t-1} + b_o)\\
\Tilde{c}_t & = \delta_{h}{(W_c x_t + V_c h_{t-1} + b_c)} \\
c_t & = i_t \odot \Tilde{c}_t + f_t \odot c_{t-1}\\
h_t & = o_t \odot \delta_{h}(c_t)
\end{split}
\end{equation}

\subsection{EDSA-Ensemble: Event Detection Sentiment Analysis Ensemble}

The Event Detection methods produce a list of event keywords. 
Each of these lists encapsulates the overall topic.
To each topic, a list with all the tweets discussing it is also attached.
By applying Sentiment Analysis models over the list of tweets belonging to a topic, then we can determine the positive or negative impact of the event over the network.

By combining both the event magnitude, i.e., the importance of the event, its lifespan, and the event's polarity, we can better understand user behavior and how the event's topic impacts and the attention given to it influences the overall user response.
Thus, we can analyze if topics that have a longer lifespan and higher magnitude influence the users in a positive or negative way.
Furthermore, by analyzing individual tweets we can determine opinion-based communities.

The proposed model EDSA-Ensemble (Event Detection Sentiment Analysis Ensemble Model) is implemented through Algorithm~\ref{alg:alg_sae}.
As input, the EDSA-Ensemble receives a set of documents \emph{D}.
The output of the model is 
(1) an \mbox{individual-sentiment} set $I_S$ that contains the individual sentiment of each user for each event obtained by each Event Detection method \textit{ED-Methods} and
(2) an \mbox{event-sentiment} set $E_S$ that contains the sentiment obtained for each event obtained by each Event Detection method \textit{ED-Methods}.
The individual and event sentiment are detecting using an ensemble of Sentiment Analysis models.
Thus, for each event $e_i \in E$ determined by an Event Detection method $ED_{method} \in$ \textit{ED-Methods}, we first compute the polarity obtained by each $SA_{models} \in$ \textit{SA-Models}.
Afterward, we compute the overall polarity of an event $e_i$  given by a $ED_{method}$ by taking the most frequent sentiment, i.e., the $Mode$, obtained by the different \textit{SA-Models}.
To improve the runtime performance of the algorithm, all the iterations over the \textit{ED-Methods} and {SA-Models} are done in parallel.

\begin{algorithm}[!ht]
\tiny
\SetKwInOut{Input}{Input}
\SetKwInOut{Output}{Output}
\Input{a document set $D$}
\Output{an individual-sentiment list $I_S$ and an event-sentiment set $E_S$}
\tcp{initialize the individual-sentiment and event-sentiment sets}
\emph{$I_S = \emptyset$}\;\label{line1}
\emph{$E_S = \emptyset$}\;\label{line2}
\tcp{for each event detection algorithm}
\ForEach{$ED_{method} \in \text{ED-Methods}$}{ \label{line3}
    \tcp{initialize the event detection method individual-sentiment and event-sentiment sets}
    \emph{$I_S^{method} = \emptyset$}\;\label{line4}
    \emph{$E_S^{method} = \emptyset$}\;\label{line5}
    \tcp{event detection text preprocessing}
    \emph{$D_C = EDTP(D, ED_{method})$}\; \label{line6}
    \tcp{extract the events}
    \emph{$E = ED(D_C, ED_{method})$}\; \label{line7}
    \ForEach{$e_i \in E$}{ \label{line8}
        \tcp{initialize the sentiment analysis model individual-sentiment and event-sentiment sets}
        \emph{$I_{EDSA} = \emptyset$}\;\label{line9}
        \emph{$E_{EDSA} = \emptyset$}\;\label{line10}
        \ForEach{$SA_{model} \in \text{SA-Models}$}{ \label{line11}
            \tcp{initialize the sentiment set $S^{(i)}$ for event $e_i$}
            \emph{$S^{(i)}=\emptyset$}\; \label{line12}
            \tcp{extract the set of documents for event $e_i$}
            \emph{$D^{(i)} = extract(e_i, D)$}\; \label{line13}
            \tcp{sentiment analysis text preprocessing}
            \emph{$D^{(i)}_C = SATP(D^{(i)}, SA_{model})$}\; \label{line14}
            \ForEach{$d \in D^{(i)}_C$}{\label{line15}
                \tcp{extract polarity for each documents $d$}
                \emph{$s = SA(d, SA_{model})$}\;\label{line16}
                \tcp{update the sentiment list}
                \emph{$S^{(i)} = S^{(i)} \cup \{s\}$}\;\label{line17}
             }
            \tcp{extract the most frequent sentiment $s^{(i)}_f$ for event $e_i$}
            \emph{$s^{(i)}_{f} = Mode(S^{(i)}, SA_{model})$}\label{line18}\;
            \tcp{update $E_{EDSA}$ with the sentiment $s^{(i)}_f$ of event $e_i$}
            \emph{$I_{EDSA} = I_{EDSA} \cup \{S^{i}\}$}\;\label{line19}
            \tcp{update $E_{EDSA}$ with the sentiment $s^{(i)}_f$ of event $e_i$}
            \emph{$E_{EDSA} = E_{EDSA} \cup \{(e_i, s^{(i)}_f)\}$}\;\label{line20}
        }
        \tcp{update $I_S^{method}$ with the most frequent sentiment}
        \emph{$I_S^{method} = Mode(I_{EDSA})$}\;\label{line21}
        \tcp{update $E_S^{method}$ with the most frequent sentiment of event $e_i$}
        \emph{$E_S^{method} = Mode(E_{EDSA})$}\;\label{line22}
    }
    \tcp{update $I_S$ with the most frequent individual-sentiment}
    \emph{$I_S = I_S \cup \{I_S^{method}\}$}\;\label{line23}
    \tcp{update $E_S$ with the most frequent sentiment of event $e_i$}
    \emph{$E_S = E_S \cup \{E_S^{method}\}$}\;\label{line24}
}
\tcp{return the $I_S$ and $E_S$ lists}
\Return{$I_S$, $E_S$}\;\label{line25}   
    
\caption{EDSA-Ensemble}\label{alg:alg_sae}
\end{algorithm}\DecMargin{1em}

The algorithm works as follows.
Firstly, we initialize the the individual-sentiment set $I_S$ (Line~\ref{line1}) and \mbox{event-sentiment} set $E_S$(Line~\ref{line2}).
This set contains the most frequent sentiment obtained by each \textit{SA-Models} of each event $e_j$ obtained by each \textit{ED-Methods}.
Secondly, we iterate over each \textit{ED-Methods} (Line~\ref{line3}).
For each $ED_{method}$, 
we initialize an event-sentiment set $E_S^{method}$ (Line~\ref{line5}), 
preprocess the text using the text preprocessing function $EDTP$ depended on the for each $ED_{method}$ and extract the preprocess corpus $D_C$ (Line~\ref{line6}), 
and extract the events for the $ED_{method}$ and store it in the set $E$ (Line~\ref{line7}).

For each event $e_i \in E$ (Line~\ref{line8}), we initialize 
(1) and individual-sentiment set $I_{EDSA}$ (Line~\ref{line9}) where we store the most frequent sentiment obtained by each of the \textit{SA-Models} and
(2) an event-sentiment set $E_{EDSA}$ (Line~\ref{line10}) where we store the most frequent sentiment obtained by each of the \textit{SA-Models} for an event (Line~\ref{line11}).
To obtain the sentiment using a model $SA_{model}$ of an event $e_i$, we
initialize an empty list  $S^{(i)}=\emptyset$ for storing the sentiment for each document belonging to the current analyzed event $e_i$ (Line~\ref{line12}).
Then, we extract the set of documents $D^{(i)} = extract(e_i, D)$ that belong to the event $e_i$  (Line~\ref{line13}) and then obtain the clean document set $D^{(i)}_C = SATP(D^{(i)})$ by applying the text preprocessing steps for the current $SA_{model}$ (Line~\ref{line14}).

For each document $d$ in set $D^{(i)}_C$, we detect the sentiment $s = SA(d)$ and update the sentiment list $S^{(i)} = S^{(i)} \cup \{s\}$ for the event $e_i$ (Lines~\ref{line15} and~\ref{line17}).
During this step, we extract individual user opinions regarding the event.
After these steps, we label the event $e_i$ with its most frequent sentiment $s^{(i)}_{f} = Mode(S^{(i)})$ (Lines~\ref{line18} and update 
(1) the individual-sentiment set $I_{EDSA} = I_{EDSA} \cup \{S^{i}\}$ (Line~\ref{line19}).
and
(2) the event-sentiment set $E_{EDSA} = E_{EDSA} \cup \{(e_i, s^{(i)}_{f})\}$ for the current $SA_{model}$ (Line~\ref{line20})

After we obtain a polarity of an event $e_i$ with each $SA_{model}$, we update 
(1) the $I_S^{method}$ with the most frequent individual sentiment $I_S^{method} = Mode(I_{EDSA})$ (Line~\ref{line21})
and
(2) the $E_S^{method}$ with the most frequent sentiment $E_S^{method} = Mode(E_{EDSA})$ for the event $e_i$ (Line~\ref{line22}).
After we obtain the ensemble results, i.e., $E_S^{method}$ and $I_S^{method}$, we update 
(1) the individual-sentiment set $I_S$ (Line~\ref{line23}) and
(2) the event-sentiment set $E_S$ (Line~\ref{line24})

When we finished the iterations over the \textit{ED-Methods}, the algorithm returns the final individual-sentient set $I_S$ and event-sentiment set $E_S$ (Line~\ref{line25}).

\section{Experimental Results}\label{ExperimentalResults}

In this section, firstly, we present the dataset used for the tree tasks: (1) Event Detection, (2) Individual Sentiment Analysis, and (3) Event Sentiment Analysis. 
Secondly, we discuss the results obtained with the event detection algorithms individually.
Thirdly, we analyze the results obtained for the individual user sentiment of events when using each classification model separately and when using EDSA-Ensemble. 
Finally, we present the result of detecting the overall sentiments of events for each individual model and the EDSA-Ensemble.

\subsection{Dataset} \label{subsec:dataset} 

We use \href{http://sentiment140.com/}{Sentiment140} for our experiments.
The dataset contains $1.6$ million annotated tweets with the labels negative, neutral, and positive.
Besides the textual content, the tweets are also timestamped with the date.
The date and the textual content are used to extract events, while the textual content and the labels are employed to build the classification models used by EDSA-Ensemble. 

To detect events and to determine the impact of text preprocessing on this task, we create three different text preprocessing pipelines: Minimal Text (MT), Pure Text (PT), and Clean Text (CT).
After preprocessing, we weight the terms to prepare the text for the event detection algorithms.
We employ the raw frequency ($f_{t_{j},d_{i}}$), i.e., the number of times a term appears in a document, to weight the words before extracting events with MABED and OLDA.
For Peaky Topics, we used the $TFIDF$ weighting scheme.
We use the entire corpus for the Event Detection experiments.
Table~\ref{tab:ED_textcleaning} presents the steps employed to clean the text used by the three different preprocessing pipelines.

\begin{table}[!htbp]
\centering
\caption{Event detection text preprocessing}
\label{tab:ED_textcleaning}
\begin{tabular}{lccc}
\toprule
\textbf{Text Preprocessing} & \textbf{MT} & \textbf{PT} & \textbf{CT}          \\
\midrule
Lowercase the text              & \checkmark        & \checkmark   & \checkmark   \\
Split the text into tokens      & \checkmark        & \checkmark   & \checkmark   \\
Remove punctuation              & \checkmark        & \checkmark   & \checkmark   \\
Remove stop words               & x                 & \checkmark   & \checkmark   \\
Extract the lemma               & x                 & x            & \checkmark   \\
\bottomrule
\end{tabular}
\end{table}

For sentiment analysis, we apply two different preprocessing pipelines: Sentiment Clean Text (SCT) and Sentiment Clean Text with Feature (SFE).
When using the SCT pipeline, the text is split into tokens and the punctuation is removed.
To preserve the sentiment polarity, we keep the case of words, duplicate letters in words, and stop words.
When using the SFE pipeline, besides splitting the text into tokens and removing punctuation, we apply feature engineering. 
First, we expand the contractions, e.g., "don't" becomes "do not".
Second, we enhance the text by combining the negation with the words that follow, i.e., "not good" becomes "not\_good".
Table~\ref{tab:SA_textcleaning} presents the steps used to create SCT and SFE.

\begin{table}[!htbp]
\centering
\caption{Sentiment analysis text preprocessing}
\label{tab:SA_textcleaning}
\begin{tabular}{lcc}
\toprule
\textbf{Text Preprocessing} & \textbf{SCT} & \textbf{SFE} \\
\midrule
Lowercase the text              & x                 & x          \\
Split the text into tokens      & \checkmark        & \checkmark \\
Remove punctuation              & \checkmark        & \checkmark \\
Remove stop words               & x                 & x          \\
Extract the lemma               & x                 & \checkmark \\
Expand contractions             & x                 & \checkmark \\
Text enhancement                & x                 & \checkmark \\
\bottomrule
\end{tabular}
\end{table}

For Sentiment Analysis, we split the corpus into three subsets to determine the scalability of the Sentiment Analysis models.
We create two subsets, i.e.,  $C_1$ and $C_2$, from the entire dataset, i.e., $C_3$, by selecting tweets at random while keeping the label distribution not to over-fit the models.
Table~\ref{tab:ds-size} presents the three datasets and the number of features for both preprocessing pipelines.

\begin{table}[!hpbt]
\centering
\caption{Datasets for Sentiment Analysis}
\label{tab:ds-size}
\begin{tabular}{lrrr}
\toprule
\textbf{Dataset}  & \textbf{Tweets} & \textbf{SCT features} & \textbf{SFE Features} \\ 
\midrule
$C_1$   & 20\,000     &  30\,651  & 31\,018 \\
$C_2$   & 500\,000    & 307\,890  & 311\,240 \\
$C_3$   & 1\,600\,233 & 684\,492  & 691\,646 \\
\bottomrule
\end{tabular}
\end{table}

After applying text preprocessing on the three subsets, i.e., $C_1$, $C_2$, and $C_3$, the words are weighted as follows.
We use the raw frequency ($f_{t_{j},d_{i}}$) for the Naïve Bayes (NB) and Ridge Classifier (RC) models.
For the Logistic Regression (LR) and Support Vector Machines (SVM), we encode the text using $TFIDF$.
We create word embeddings using Word2Vec to encode the textual data and train the LSTM model.
We use transformer embeddings, i.e., BERT, RoBERTa, and XLM-RoBERTa, to train the Perceptron model.

\subsection{Event Detection}

For this first set of experiments, we extract $50$ events for the time window between April and July 2009
Each extracted event is described by the top-10 most relevant keywords.
To determine which prepossessing strategy is better suited to extract clear and human-readable events, we did experiments with all three text preprocessing pipelines, i.e., MT, PT, and CT.

For the first set of experiments, we observe that MABED manages to extract human-readable events when employing CT that spread over a longer period of time with a larger number of tweets discussing that topic, i.e., the magnitude value is big.
When using MT or PT, the lifespan of an event is shortened and the number of tweets discussing the event is small and dispersed, as the magnitude is small.
Also, the events become more specific.
Table~\ref{tab:MABEDMethod} presents an event sample detected using MABED when employing CT.

\begin{table}[!htbp]
\centering
\caption{Top-10 events by magnitude detected with MABED using CT}
\label{tab:MABEDMethod}
\resizebox{\columnwidth}{!}{%
\begin{tabular}{rlll}
\toprule
\textbf{Magnitude} & \textbf{Star Date} & \textbf{End Date} & \textbf{Topic}          \\
\midrule
82\,736 & 2009-04-06-22-19-49 & 2009-06-25-10-26-23 & night nt tomorrow hate sad na cry feeling wan ca \\
68\,278 & 2009-04-06-22-19-49 & 2009-06-25-10-26-23 & na nt sad sleep ca quot call hate feeling \\
67\,403 & 2009-04-06-22-19-49 & 2009-06-25-10-28-09 & nt sleep wa na morning gon early class getting school \\
59\,412 & 2009-04-06-22-20-19 & 2009-06-25-10-24-29 & na bed night class nt tomorrow late luck fun wa \\
57\,494 & 2009-04-06-22-19-49 & 2009-06-25-10-26-23 & tomorrow home ca nt soon happy late \\
56\,798 & 2009-04-06-22-20-19 & 2009-06-25-10-28-26 & nt night feel show wa hour fun doe \\
52\,305 & 2009-04-06-22-20-19 & 2009-06-25-10-26-21 & na phone wa class quot girl nt star wish watch \\
49\,566 & 2009-04-06-22-19-49 & 2009-06-25-10-28-09 & nt wait find sad early school wan na sleep \\
48\,770 & 2009-04-06-22-21-21 & 2009-06-25-10-26-04 & gon nt wan bed sad miss sleep tomorrow omg night \\
47\,953 & 2009-04-06-22-20-19 & 2009-06-25-10-28-25 & look quot nt na wa hate feeling soo family show \\
\bottomrule
\end{tabular}
}
\end{table}

Similar to the MABED results, we observe that the events extracted by OLDA when using CT are human-readable, spread over a longer period of time, and the users are more engaged in discussing the event.
As OLDA uses the entire vocabulary to summarize a topic picking words using Gibbs sampling, the keywords are more related to each other.
Thus, when using OLDA together with the CT text prepossessing pipeline, the topics are more generic and cover more tweets.
Furthermore, when using the MT or PT, we observe the same pattern emerging as in the case of MABED: the topics have lower magnitude, i.e., the time window is smaller and the number of tweets discussing a topic decrease, while the events become more specific. 
Table~\ref{tab:OLDAMethod} presents a sample of topics detected using OLDA when employing CT.

\begin{table}[!htbp]
\centering
\caption{Top-10 events by magnitude detected with OLDA using CT}
\label{tab:OLDAMethod}
\resizebox{\columnwidth}{!}{%
\begin{tabular}{rlll}
\toprule
\textbf{Magnitude} & \textbf{Star Date} & \textbf{End Date} & \textbf{Topic}          \\
\midrule
107\,902 & 2009-04-06-22-19-53 & 2009-06-25-10-28-26 & day work today going back tomorrow home go morning time \\
105\,031 & 2009-04-06-22-20-19 & 2009-06-25-10-28-28 & going wish could yes wa would nt get said go \\
 94\,126 & 2009-04-06-22-19-49 & 2009-06-25-10-28-28 & nt ca wait get doe hurt suck wo make see \\
 68\,211 & 2009-04-06-22-19-53 & 2009-06-25-10-26-29 & na gon wan go tired bed see http time im \\
 66\,763 & 2009-04-06-22-22-52 & 2009-06-25-10-26-20 & night last better hope good well feeling everyone get oh \\
 65\,412 & 2009-04-06-22-19-49 & 2009-06-25-10-28-28 & wa sick today nt one think thought fuck made many \\
 61\,243 & 2009-04-06-22-20-37 & 2009-06-25-10-28-26 & good happy morning hey birthday thank best luck day wa \\
 60\,642 & 2009-04-06-22-19-53 & 2009-06-25-10-28-09 & want thanks tweet back go get ok call time really \\
 55\,651 & 2009-04-06-22-20-20 & 2009-06-25-10-26-38 & wa miss week home next time back come year fun \\
 48\,926 & 2009-04-06-22-19-57 & 2009-06-25-10-28-30 & like feel look watching sorry looking sound good know cool \\
\bottomrule
\end{tabular}
}
\end{table}

Peaky Topic manages to extract accurate events and bursty topics but the topic of readability is influenced by the preprocessing strategy we employed.
Due to the nature of this method, the equally distribute time-frames and $TFIDF$, the events spread over all the time intervals and have high magnitude.
Table~\ref{tab:PTMethod} presents a sample of topics detected using Peaky Topics when employing CT.

\begin{table}[!htbp]
\centering
\caption{Top-10 events by magnitude detected with Peaky Topics using CT}
\label{tab:PTMethod}
\resizebox{\columnwidth}{!}{%
\begin{tabular}{rlll}
\toprule
\textbf{Magnitude} & \textbf{Star Date} & \textbf{End Date} & \textbf{Topic}          \\
\midrule
93\,973 & 2009-04-06 22:19:45 & 2009-06-25 10:27:58 & get wa work got like know im http hope home \\
82\,960 & 2009-04-06 22:21:07 & 2009-06-25 10:28:02 & home one got go work haha lol like last know \\
77\,458 & 2009-04-06 22:20:20 & 2009-06-25 10:27:58 & home really tomorrow going wa would lol like last know \\
71\,859 & 2009-04-06 22:19:45 & 2009-06-25 10:28:00 & would got love lol like last know im http home \\
67\,749 & 2009-04-06 22:19:45 & 2009-06-25 10:28:31 & wish ha work got lol like last know im http \\
66\,630 & 2009-04-06 22:21:49 & 2009-06-25 10:25:52 & quot still work haha love lol like last know im \\
66\,562 & 2009-04-06 22:22:47 & 2009-06-25 10:28:26 & going would haha make love lol like last know im \\
64\,854 & 2009-04-06 22:22:48 & 2009-06-25 10:28:26 & http would haha make love lol like last know im \\
63\,870 & 2009-04-06 22:22:48 & 2009-06-25 10:28:26 & would ha make love lol like last know im http \\
63\,109 & 2009-04-06 22:22:48 & 2009-06-25 10:26:22 & im would great love lol like know http hope home \\
\bottomrule
\end{tabular}
}
\end{table}

For a quick comparison of how those Event Detection methods behave, we have extracted 5 events for each method and used the top-10 most relevant terms (Table~\ref{tab:EDMethods}). 

\begin{table}[!htbp]
\centering
\caption{Comparison between Top 5 topics for each Event Detection Method}
\label{tab:EDMethods}
\resizebox{\columnwidth}{!}{%
\begin{tabular}{lrccl}
\toprule
\textbf{Method} & \textbf{Magnitude} & \textbf{Start Date} & \textbf{End Date}  & \textbf{Topic}            \\ 
\midrule

MABED & 82\,736 & 2009-04-06-22-19-49 & 2009-06-25-10-26-23 & night nt tomorrow hate sad na cry feeling wan ca   \\
MABED & 68\,278 & 2009-04-06-22-19-49 & 2009-06-25-10-26-23 & na nt sad sleep ca quot call hate feeling     \\
MABED & 67\,403 & 2009-04-06-22-19-49 & 2009-06-25-10-28-09 & nt sleep wa na morning gon early class getting school  \\
MABED & 59\,412 & 2009-04-06-22-20-19 & 2009-06-25-10-24-29 & na bed night class nt tomorrow late luck fun wa   \\
MABED & 52\,305 & 2009-04-06-22-20-19 & 2009-06-25-10-26-21 & na phone wa class quot girl nt star wish watch   \\

\midrule

OLDA & 107\,902 & 2009-04-06-22-19-53 & 2009-06-25-10-28-26 & day work today going back tomorrow home go morning time \\
OLDA & 105\,031 & 2009-04-06-22-20-19 & 2009-06-25-10-28-28 & going wish could yes wa would nt get said go    \\
OLDA & 66\,763 & 2009-04-06-22-22-52 & 2009-06-25-10-26-20 & night last better hope good well feeling everyone get oh \\
OLDA & 65\,412 & 2009-04-06-22-19-49 & 2009-06-25-10-28-28 & wa sick today nt one think thought fuck made many   \\
OLDA & 36\,669 & 2009-04-06-22-21-39 & 2009-06-25-10-27-58 & getting today ready http sad get hot still heart thing \\

\midrule

Peaky Topics & 93\,973 & 2009-04-06 22:19:45 & 2009-06-25 10:27:58 & get wa work got like know im http hope home    \\
Peaky Topics & 77\,458 & 2009-04-06 22:20:20 & 2009-06-25 10:27:58 & home really tomorrow going wa would lol like last know \\
Peaky Topics & 66\,630 & 2009-04-06 22:21:49 & 2009-06-25 10:25:52 & quot still work haha love lol like last know im   \\
Peaky Topics & 63\,109 & 2009-04-06 22:22:48 & 2009-06-25 10:26:22 & im would great love lol like know http hope home   \\
Peaky Topics & 55\,872 & 2009-04-06 22:21:39 & 2009-06-25 10:26:35 & oh tonight na like work great know im http hope   \\
\bottomrule
\end{tabular}
}
\end{table}

When analyzing the most representative topic keywords using the text preprocessing CT, we can observe that MABED, OLDA, and Peaky Topics manage to detect different emerging events, although some have common keywords (Table~\ref{tab:MABED-OLDA}).
MABED provides better results, as it also uses fewer words.
Peaky Topics does not detect the same results as the other two algorithms, as it uses another type of topic detection approach.
Although the magnitudes of the events are somehow comparable, meaning they are all close to each other, the relevant terms are different.
Because of the type of the model, no intervals overlap and for that, it is hard to compare its results to the other 2 methods based on time windows.

\begin{table}[!hpbt]
\caption{Events with common keywords}
\label{tab:MABED-OLDA}
\resizebox{\columnwidth}{!}{%
\begin{tabular}{llll}
\toprule
\textbf{Algorithm} & \textbf{Term Weight}     & \textbf{Sim \#} & \textbf{Event Keywords}           \\
\midrule
\multirow{4}{*}{\textbf{MABED}}  & \multirow{4}{*}{$\bm{TFIDF}$} & 1    & wa night nt th quot ate home wow amazing       \\
          & & 2    & nt feel hope twitter read book fan ah       \\
          & & 3    & tomorrow happy quot hey hope thought sleeping wa     \\ 
          & & 4    & nt wish amp thought watch tomorrow hour bed meeting bad   \\ 
\midrule
\multirow{4}{*}{\textbf{OLDA}}  & \multirow{4}{*}{$\bm{f_{t_{j}, d_{j}}}$}   & 1    & back home come going la tomorrow go came heading write   \\
          & & 2    & twitter friend best leaving wont bye facebook know lol let  \\
          & & 3    & day another today long tomorrow dad beautiful summer going start \\
          & & 4    & work sleep hour tired done early get go still woke    \\
\midrule
\multirow{4}{*}{\textbf{Peaky Topics}}  & \multirow{4}{*}{$\bm{TFIDF}$} & 1    & home one got go work haha lol like last know      \\
          & & 2    & twitter today would great love lol like last know im    \\
          & & 3    & home really tomorrow going wa would lol like last know   \\
          & & 4    & wish ha work got lol like last know im http      \\
          
\bottomrule
\end{tabular}
}
\end{table}

As a conclusion to the comparison of the 3 methods, we can say the following: 
\begin{itemize}
 \item MABED and OLDA are good for long-term topic analysis, as they take into consideration the evolution in time.
 \item Peaky Topics is good for taking immediate decisions to moderate content, as it may be harmful to the network.
 \item Because of the nature of the summarizing process, neither method produces easily understandable content.
 \item In terms of magnitude divided by the amount of time, all methods are comparable. 
 The same can be deducted for MABED and OLDA when it comes to the total amount of time.
\end{itemize}

\subsection{Individual Sentiment Analysis}

For the Individual Sentiment Analysis tasks, we use Naïve Bayes (NB), Ridge Classification (RC), Logistic Regression (LR), and Support Vector Machine (SVM) together with SCT and SFE text preprocessing strategies.
For the Long-Short Term Memory (LSTM), we used the Word2Vec embeddings extracted for both preprocessing strategies.
We employ the Accuracy, Precision, and Recall metrics for evaluating the results, and K-Folds for the evaluation scores.
We also compare the results obtained by each individual model with the results obtained by the proposed EDSA-Ensemble model for individual user opinion.
Table~\ref{tab:results_sa} presents the results for this task.

The first set of experiments for Sentiment Analysis uses Naïve Bayes (NB) together with SCT.
The evaluation scores improve with the size of the text. We considered these scores the baseline for our experiments.
When using NB with SFE, the accuracy and recall slightly increase while precision decreases regardless of the dataset size.

When using LR with SCT, the evaluation scores improve with the size of the text.
We observe that the accuracy is slightly lower than the one obtained with NB.
When using LR with SFE, the evaluation scores worsen for the small subset of documents $C_1$, but they improve with the scale. Overall, the scores improve compared to the results obtained with NB.

When using the SCT with Ridge Classifier (RC), we obtain better scores than for LR and NB for all the evaluation metrics. 
The performance of RC worsens when using the SFE. 
This is a direct impact of using the $TFIDF$ instead of the raw term frequency.

\begin{table}[!ht]
\caption{Sentiment Analysis Results}~\label{tab:results_sa}
\resizebox{\columnwidth}{!}{%
\begin{tabular}{cllrrrrrr}
\toprule
\multirow{2}{*}{\textbf{Dataset}}            & \multirow{2}{*}{\textbf{Algorithm}} &  \multirow{2}{*}{\textbf{Term Weight}}    & \multicolumn{3}{c}{\textbf{SCT}}  & \multicolumn{3}{c}{\textbf{SFE}}  \\
    &  &  & \textbf{Accuracy} & \textbf{Precision} & \textbf{Recall} & \textbf{Accuracy} & \textbf{Precision} & \textbf{Recall} \\
\midrule
\multirow{9}{*}{$\bm{C_1}$}  & \textbf{NB} & $\bm{f_{{t_j}, d_{i}}}$  & 0.761    & 0.706    & 0.841   & 0.754   & 0.694    & 0.852 \\
                             & \textbf{LR}   & $\bm{TFIDF}$ & 0.756    & 0.703    & 0.766   & 0.752   & 0.690    & 0.752 \\
                             & \textbf{RC}  & $\bm{f_{{t_j}, d_{i}}}$  & 0.767    & 0.710    & 0.852   & 0.753   & 0.690    & 0.861 \\
                             & \textbf{SVM}  & $\bm{TFIDF}$  & 0.729    & 0.707    & 0.772   & 0.748   & 0.725    & 0.796 \\
                             & \textbf{LSTM} & \textbf{Word2Vec}  & 0.930    & 0.931    & 0.930   & 0.923   & 0.922    & 0.923 \\
                             & \textbf{Perceptron} & \textbf{BERT}  & 0.828    & 0.828    & 0.828   & 0.776   & 0.776    & 0.776 \\
                             & \textbf{Perceptron} & \textbf{RoBERTa}  & 0.866    & 0.866    & 0.866   & 0.777   & 0.777    & 0.777 \\ 
                             & \textbf{Perceptron} & \textbf{XLM-RoBERTa} & 0.828    & 0.828    & 0.828   & 0.776   & 0.776    & 0.776 \\ \cline{2-9}
                             &  \multicolumn{2}{c}{\textbf{EDSA Ensemble}} & \textbf{0.949} & \textbf{0.950} & \textbf{0.948} & \textbf{0.940} & \textbf{0.941} & \textbf{0.940} \\
\midrule
\multirow{9}{*}{$\bm{C_2}$}  & \textbf{NB} & $\bm{f_{{t_j}, d_{i}}}$   & 0.794    & 0.752    & 0.857   & 0.788   & 0.738    & 0.867 \\
                             & \textbf{LR} & $\bm{TFIDF}$   & 0.790    & 0.727    & 0.802   & 0.794   & 0.730    & 0.807 \\
                             & \textbf{RC} & $\bm{f_{{t_j}, d_{i}}}$   & 0.810    & 0.772    & 0.862   & 0.794   & 0.747    & 0.865 \\
                             & \textbf{SVM} & $\bm{TFIDF}$   & 0.778    & 0.799    & 0.742   & 0.781   & 0.763    & 0.815 \\
                             & \textbf{LSTM} & \textbf{Word2Vec}  & 0.855    & 0.859    & 0.855   & 0.850   & 0.851    & 0.850 \\ 
                             & \textbf{Perceptron} & \textbf{BERT}  & 0.856    & 0.856    & 0.856   & 0.800   & 0.800    & 0.800 \\
                             & \textbf{Perceptron} & \textbf{RoBERTa}  & 0.887    & 0.887    & 0.887   & 0.809   & 0.809    & 0.809 \\
                             & \textbf{Perceptron} & \textbf{XLM-RoBERTa} & 0.863    & 0.863    & 0.863   & 0.807   & 0.807    & 0.807 \\ \cline{2-9}
                             &  \multicolumn{2}{c}{\textbf{EDSA Ensemble}} & \textbf{0.901} & \textbf{0.900} & \textbf{0.902} & \textbf{0.877} & \textbf{0.878} & \textbf{0.877} \\
\midrule
\multirow{9}{*}{$\bm{C_3}$}  & \textbf{NB} & $\bm{f_{{t_j}, d_{i}}}$   & 0.796    & 0.749    & 0.863   & 0.786   & 0.731    & 0.873 \\
                             & \textbf{LR} & $\bm{TFIDF}$   & 0.800    & 0.739    & 0.810   & 0.804   & 0.742    & 0.815 \\ 
                             & \textbf{RC} & $\bm{f_{{t_j}, d_{i}}}$   & 0.814    & 0.774    & 0.865   & 0.794   & 0.745    & 0.866 \\ 
                             & \textbf{SVM} & $\bm{TFIDF}$   & 0.803    & 0.795    & 0.820   & 0.806   & 0.795    & 0.825 \\ 
                             & \textbf{LSTM} & \textbf{Word2Vec}  & 0.824    & 0.827    & 0.824   & 0.835   & 0.835    & 0.835 \\ 
                             & \textbf{Perceptron} & \textbf{BERT}  & 0.863    & 0.863    & 0.863   & 0.807   & 0.807    & 0.807 \\ 
                             & \textbf{Perceptron} & \textbf{RoBERTa}  & 0.892    & 0.892    & 0.892   & 0.817   & 0.817    & 0.817 \\ 
                             & \textbf{Perceptron} & \textbf{XLM-RoBERTa} & 0.870    & 0.870    & 0.870   & 0.814   & 0.814    & 0.814 \\ \cline{2-9}
                             &  \multicolumn{2}{c}{\textbf{EDSA Ensemble}} & \textbf{0.912} & \textbf{0.913} & \textbf{0.910} & \textbf{0.872} & \textbf{0.873} & \textbf{0.872} \\
\bottomrule
\end{tabular}
}
\end{table}

For the SVM approach, we use a \textbf{linear kernel} as the classes are \textbf{balanced}. 
We set the penalty parameter of the error term $C=0.1$, value obtained after hyperparameter tuning.
When using SVM with SCT, we observe that the evaluation scores improve with the increasing number of used features, with an overall increase of almost $\sim 0.5$.
For SVM with SFE, the results show the same ascending trend of the score improvement w.r.t. the scale of the dataset. 
As in the case of LR, we can observe that the improvement in the evaluation scores is minimal w.r.t. the text preprocessing strategy we employed, i.e., overall for all the measures $\sim 0.01$. Furthermore, the overall score improvement for all the measures w.r.t. the scale is also minimal, i.e., $\sim 0.05$. 
Thus we can conclude once more that the number of features does not impact any of the measures. 

When using LSTM with Word2Vec, although we obtain the best results regarding the accuracy of the Sentiment Analysis process, this comes at the cost of high runtime and resource utilization. 
We can observe the following:
\begin{itemize}
 \item For the same configuration, the more data the better.
 \item Based on the number of features, the same configuration (layers and numbers of neurons) produces different results without scaling the training data.
\end{itemize}

For LSTM, the results for $C_1$ have a peak of about $0.93$ accuracy for SCT and $0.92$ for SFE.
The accuracy decreases with the size of the corpus.
For the corpus of a smaller size, LSTM overfits.
The experiment for $C_3$ needs to be run on more powerful hardware such as dedicated HPC GPUs as a batch takes around 24 hours.

By analyzing the experimental results, we can observe that the improvement in the evaluation scores is minimal w.r.t. our employed text preprocessing strategy, i.e., overall for all the measures $\sim 0.004$. 
Furthermore, the scores improvement w.r.t. the scale is also minimal, i.e., overall for all the measures $\sim 0.05$. 
Thus, we can conclude that the number of features does not impact any of the measures. 

When using the $TFIDF$ weighting scheme, we observe that the results slightly worsen if we employ SFE instead of SCT.
For the overall tests, with parameters tuning and cross-validation, we can see that the RC performs better than LR but worse than SVM.

SVM provides better evaluation scores than all the other algorithms that use the Bag-of-Words model for text preprocessing with regard to the number of features, but the performance improvement is minimal, i.e., $\sim 0.05$.
The best results were given by the LSTM at the cost of processing power.
Using SFE with raw frequency weighting schema improves the evaluation scores for LR and SVM. 
When using $TFIDF$ and Word2Vec, the evaluation scores for NB, RC, and LSTM slightly worsen. 

The EDSA-Ensemble model uses all eight individual models to determine the polarity of tweets.
The proposed model uses a voting process to determine for each tweet which polarity was the most frequently determined by each individual model.
When using the EDSA-Ensemble model, all the evaluation metrics are increased. 
This shows that individual model miss-classification is mitigated through the voting process.

From a runtime perspective when training, LR is much faster than SVM (Table~\ref{tab:results_sa_time}), as the model construction takes minutes (m) for LR, while for SVM it takes days (D). 
We can conclude that the LR algorithm is the best choice for constructing the Sentiment Analysis model.
Although the SVM achieves a very small increase in evaluation score over LR, the waiting time for building the model is not feasible. 
The peak performance comes from LSTM which takes a lot of time and resources to train.
The EDSA-Ensemble model runtime for training performance is equal to the slowest model, i.e., SVM, as it trains all the Sentiment analysis models in parallel.

\begin{table}[!hpbt]
\centering
\caption{Runtime comparison}
\label{tab:results_sa_time}
\begin{tabular}{llrrrrrr}
\toprule
\multirow{2}{*}{\textbf{Algorithm}} &  \multirow{2}{*}{\textbf{Term Weight}} & \multicolumn{3}{c}{\textbf{SCT}} & \multicolumn{3}{c}{\textbf{SFE}} \\ 
 &  & \multicolumn{1}{c}{$\bm{C_1}$} & \multicolumn{1}{c}{$\bm{C_2}$} & \multicolumn{1}{c}{$\bm{C_3}$} & \multicolumn{1}{c}{$\bm{C_1}$} & \multicolumn{1}{c}{$\bm{C_2}$} & \multicolumn{1}{c}{$\bm{C_3}$} \\ 
\midrule
\textbf{NB}   & $\bm{f_{{t_j}, d_{i}}}$    &  3 m & 5 m & 17 m  & 2 m & 4.33 m & 9 m \\ 
\textbf{LR}   & $\bm{TFIDF}$    &  2 m & 5 m & 20 m  & 1.33 m & 4 m & 11 m \\ 
\textbf{RC}   & $\bm{f_{{t_j}, d_{i}}}$    &  3 m & 5 m & 20 m  & 2 m & 4 m & 10.5 m \\ 
\textbf{SVM}  & $\bm{TFIDF}$    &  2 D & 32 D & 65 D  & 1 D & 4 D & 62 D \\ 
\textbf{LSTM} & \textbf{Word2Vec}    &  1 D & 15 D & 21 D  & 1 D & 13 D & 18 D \\ 
\textbf{Perceptron} & \textbf{BERT}         & 31 m & 6.4 h & 1.11 D & 31 m & 5.85 h & 2 D \\ 
\textbf{Perceptron} & \textbf{RoBERTa}      & 31 m & 6.7 h & 1.12 D & 32 m & 5.84 h & 2 D \\ 
\textbf{Perceptron} & \textbf{XLM-RoBERTa}  & 32 m & 7 h & 1.25 D & 33 m & 6.21 h & 2 D \\ 
\bottomrule
\end{tabular}
\end{table}

\subsection{Event Sentiment Analysis}

To analyze the accuracy of EDSA-Ensemble, we run the Event Detection tasks on the entire dataset.
For these experiments, we use the CT text preprocessing strategy, discussed in subsection~\ref{subsec:dataset}. 
After this step is completed, we determine the subsets of tweets belonging to each event, on which we apply the SFE text preprocessing strategy.
For each subset of processed tweets, we use the proposed Sentiment Analysis algorithms and compute the evaluation scores. 
The results are presented in Table~\ref{tab:SAE_results}. 
Regardless of the employed Event Detection method, the overall best scores for Sentiment Analysis of events are obtained when using the proposed EDSA-Ensemble.

\begin{table}[!ht]
\centering
\caption{Average scores for Sentiment Analysis of Events}
\label{tab:SAE_results}
\resizebox{\columnwidth}{!}{%
\begin{tabular}{llllrrr}
\toprule
\multicolumn{1}{c}{\begin{tabular}[c]{@{}c@{}}\textbf{Event Detection}\\ \textbf{Model} \end{tabular}} 
& 
\multicolumn{1}{c}{\begin{tabular}[c]{@{}c@{}}\textbf{Event Detection}\\ \textbf{Term Weight} \end{tabular}}
& 
\multicolumn{1}{c}{\begin{tabular}[c]{@{}c@{}}\textbf{Sentiment Analysis} \\ \textbf{Model} \end{tabular}} 

& 
\multicolumn{1}{c}{\begin{tabular}[c]{@{}c@{}}\textbf{Sentiment Analysis}\\ \textbf{Term Weight} \end{tabular}}
&
\textbf{Accuracy} & \textbf{Precision} & \textbf{Recall} \\  
\midrule

\multirow{9}{*}{\textbf{MABED}} & \multirow{8}{*}{$\bm{TFIDF}$} 
     & \textbf{NB}    & $\bm{f_{{t_j}, d_{i}}}$ & 0.81 & 0.73 & 0.69 \\  
     & & \textbf{LR}    & $\bm{TFIDF}$ & 0.80 & 0.65 & 0.70 \\  
     & & \textbf{RC}    & $\bm{f_{{t_j}, d_{i}}}$ & 0.80 & 0.80 & 0.54 \\  
     & & \textbf{SVM}   & $\bm{TFIDF}$ & 0.79 & 0.64 & 0.66 \\  
     & & \textbf{LSTM}  & \textbf{Word2Vec} & 0.88 & 0.88 & 0.88 \\  
     & & \textbf{Perceptron} & \textbf{BERT}  & 0.95 & 0.95 & 0.95 \\  
     & & \textbf{Perceptron} & \textbf{RoBERTa}  & 0.91 & 0.92 & 0.91 \\  
     & & \textbf{Perceptron} & \textbf{XLM-RoBERTa} & 0.91 & 0.92 & 0.91 \\ \cline{3-7}
     & & \multicolumn{2}{c}{\textbf{EDSA Ensemble}} & \textbf{0.97} & \textbf{0.95} & \textbf{0.96}  \\

\midrule

\multirow{9}{*}{\textbf{OLDA}} & \multirow{8}{*}{$\bm{f_{t_{j}, d_{i}}}$}
     & \textbf{NB}   & $\bm{f_{{t_j}, d_{i}}}$ & 0.82 & 0.81 & 0.85 \\  
     & & \textbf{LR}   & $\bm{TFIDF}$ & 0.80 & 0.75 & 0.85 \\  
     & & \textbf{RC}   & $\bm{f_{{t_j}, d_{i}}}$ & 0.82 & 0.84 & 0.80 \\  
     & & \textbf{SVM}   & $\bm{TFIDF}$ & 0.78 & 0.74 & 0.83 \\  
     & & \textbf{LSTM}  & \textbf{Word2Vec} & 0.88 & 0.88 & 0.88 \\  
     & & \textbf{Perceptron} & \textbf{BERT}  & 0.95 & 0.95 & 0.95 \\  
     & & \textbf{Perceptron} & \textbf{RoBERTa}  & 0.91 & 0.92 & 0.91 \\  
     & & \textbf{Perceptron} & \textbf{XLM-RoBERTa} & 0.91 & 0.92 & 0.91 \\ \cline{3-7}
     & & \multicolumn{2}{c}{\textbf{EDSA Ensemble}} & \textbf{0.96} & \textbf{0.97} & \textbf{0.96}  \\ 
     
\midrule

\multirow{9}{*}{\textbf{Peaky Topics}} & \multirow{8}{*}{$\bm{TFIDF}$}
     & \textbf{NB}   & $\bm{f_{{t_j}, d_{i}}}$ & 0.80 & 0.80 & 0.80 \\  
     & & \textbf{LR}   & $\bm{TFIDF}$ & 0.77 & 0.71 & 0.82 \\  
     & & \textbf{RC}   & $\bm{f_{{t_j}, d_{i}}}$ & 0.80 & 0.82 & 0.75 \\  
     & & \textbf{SVM}   & $\bm{TFIDF}$ & 0.75 & 0.70 & 0.79 \\  
     & & \textbf{LSTM}  & \textbf{Word2Vec} & 0.88 & 0.88 & 0.88 \\  
     & & \textbf{Perceptron} & \textbf{BERT}  & 0.95 & 0.95 & 0.95 \\  
     & & \textbf{Perceptron} & \textbf{RoBERTa}  & 0.91 & 0.92 & 0.91 \\  
     & & \textbf{Perceptron} & \textbf{XLM-RoBERTa} & 0.91 & 0.92 & 0.91 \\ \cline{3-7}
     & & \multicolumn{2}{c}{\textbf{EDSA Ensemble}} & \textbf{0.97} & \textbf{0.97} & \textbf{0.97}  \\ 
\midrule

\end{tabular}
}
\end{table}

To better understand this gap between the scores, we analyze the results individually. Thus, for each event detected in the previous experiment, we extract the topics and label them with the most recurring sentiment. 
We also extract the most recurring sentiment detected by the Sentiment Analysis algorithms for each event. 
Tables~\ref{tab:MABED_SAE}, \ref{tab:OLDA_SAE}, and~\ref{tab:PT_SAE} present a subset of the sentiment detected for each event, i.e., positive (+) and negative (-).
The analysis of the results shows that EDSA-Ensemble is accurate and is correctly identifying the sentiment for each event, even though the individual models of the proposed ensemble are not always accurate in their prediction.

\begin{table}[!hpbt]
\centering
\caption{Sentiment Analysis of Events detected with MABED}
\label{tab:MABED_SAE}
\resizebox{\columnwidth}{!}{%
\begin{tabular}{lcccccccccc}
\toprule
\multirow{2}{*}{\textbf{MABED extracted topics}}
&  
\multirow{2}{*}{
\begin{tabular}[c]{@{}c@{}}\textbf{True}\\ \textbf{Label}\end{tabular}
}
& \textbf{NB}             & \textbf{LR}  & \textbf{RC}             & \textbf{SVM} & \textbf{LSTM}     & \textbf{Perceptron} & \textbf{Perceptron} & \textbf{Perceptron} & \multirow{2}{*}{
\begin{tabular}[c]{@{}c@{}}\textbf{EDSA}\\ \textbf{Ensemble}\end{tabular}
} \\ 
                                                                              
&  & $\bm{f_{{t_j}, d_{i}}}$ & $\bm{TFIDF}$ & $\bm{f_{{t_j}, d_{i}}}$ & $\bm{TFIDF}$ & \textbf{Word2Vec} & \textbf{BERT}       & \textbf{RoBERTa}    & \textbf{XLM-RoBERTa} \\

\midrule

nt sleep wa na morning gon early class getting school   & -                   & -                       & -            & -                       & -            & -                 & -                   & -                   & - & -                    \\ 
movie look read wa song watching getting                & +                   & +                       & +            & +                       & +            & +                 & +                   & +                   & + & +                   \\ 
phone wa miss ah finally quot doe lt facebook amp       & -                   & -                       & -            & -                       & -            & -                 & +                   & +                   & - & -                   \\ 
nt night feel show wa hour fun doe                      & -                   & -                       & -            & -                       & -            & -                 & +                   & +                   & + & -                   \\ 
course saw party quot lt dude look em hey               & +                   & +                       & +            & +                       & +            & +                 & +                   & +                   & - & +                \\ 
\bottomrule
\end{tabular}
}
\end{table}

\begin{table}[!hpbt]
\centering
\caption{Sentiment Analysis of Events detected with OLDA}
\label{tab:OLDA_SAE}
\resizebox{\columnwidth}{!}{%
\begin{tabular}{lccccccccccc}
\toprule
\multirow{2}{*}{\textbf{OLDA extracted topics}}
&  
\multirow{2}{*}{
\begin{tabular}[c]{@{}c@{}}\textbf{True}\\ \textbf{Label}\end{tabular}
}
& \textbf{NB}             & \textbf{LR}  & \textbf{RC}             & \textbf{SVM} & \textbf{LSTM}     & \textbf{Perceptron} & \textbf{Perceptron} & \textbf{Perceptron} & \multirow{2}{*}{
\begin{tabular}[c]{@{}c@{}}\textbf{EDSA}\\ \textbf{Ensemble}\end{tabular}
} \\  
                                                                              
&  & $\bm{f_{{t_j}, d_{i}}}$ & $\bm{TFIDF}$ & $\bm{f_{{t_j}, d_{i}}}$ & $\bm{TFIDF}$ & \textbf{Word2Vec} & \textbf{BERT}       & \textbf{RoBERTa}    & \textbf{XLM-RoBERTa} \\

\midrule

hurt head welcome eye pain broke coffee hand foot leg   & -                   & -                       & -            & -                       & -            & -                 & -                   & -                   & - & -                    \\

thank hot god boy stuck annoying yo sold much ff        & +                   & +                       & +            & +                       & +            & +                 & -                   & +                   & + & +                    \\
finally sunday rest sleeping smile afternoon lazy power & +                   & +                       & +            & -                       & +            & +                 & +                   & +                   & + & +                    \\
news yea lol body exciting http freaking haha oh chris  & +                   & +                       & +            & +                       & +            & +                 & +                   & +                   & + & +                    \\
hurt head welcome eye pain broke coffee hand foot leg   & -                   & -                       & -            & -                       & -            & -                 & -                   & -                   & - & -                    \\
\bottomrule
\end{tabular}
}
\end{table}

\begin{table}[!hpbt]
\centering
\caption{Sentiment Analysis of Events detected with Peaky Topics}
\label{tab:PT_SAE}
\resizebox{\columnwidth}{!}{%
\begin{tabular}{lcccccccccc}
\toprule
\multirow{2}{*}{\textbf{Peaky Topics extracted topics}}
&  
\multirow{2}{*}{
\begin{tabular}[c]{@{}c@{}}\textbf{True}\\ \textbf{Label}\end{tabular}
}
& \textbf{NB}             & \textbf{LR}  & \textbf{RC}             & \textbf{SVM} & \textbf{LSTM}     & \textbf{Perceptron} & \textbf{Perceptron} & \textbf{Perceptron} & \multirow{2}{*}{
\begin{tabular}[c]{@{}c@{}}\textbf{EDSA}\\ \textbf{Ensemble}\end{tabular}
} \\ 
                                                                              
&  & $\bm{f_{{t_j}, d_{i}}}$ & $\bm{TFIDF}$ & $\bm{f_{{t_j}, d_{i}}}$ & $\bm{TFIDF}$ & \textbf{Word2Vec} & \textbf{BERT}       & \textbf{RoBERTa}    & \textbf{XLM-RoBERTa} \\

\midrule

http would haha make love lol like last know im         & +                   & +                       & +            & +                       & +            & +                 & -                   & -                   & + & +                    \\ 
ha http home make love look lol like last know          & +                   & +                       & +            & +                       & +            & +                 & +                   & +                   & + & +                    \\ 
http work great like last know im hope home happy       & -                   & -                       & -            & -                       & +            & -                 & +                   & +                   & + & -                    \\ 
would ha make love lol like last know im http           & +                   & +                       & +            & +                       & +            & +                 & -                   & -                   & + & +                    \\ 
would got love lol like last know im http home          & +                   & +                       & -            & +                       & +            & +                 & -                   & +                   & - & +                    \\
\bottomrule
\end{tabular}
}
\end{table}

\section{Discussion}\label{discussion}

In this section, we discuss what we have found while developing our ensemble solution and analyze the results.
First, we discuss on the topic of event detection, comparing our results with the literature.
Second, we analyze the benchmarks on the sentiment analysis task and discuss performance vs. resources.
Lastly, we compare all the versions of the pipeline with regard to real-time usability and the resources required for that.

\subsection{Event Detection}

We chose three ED methods for our solution, namely MABED, OLDA, and Peaky Topics. 
These methods have different complexities of code and produce topics that do not behave the same:
\begin{itemize}
    \item MABED produces events with medium to large magnitude with 4 of them being close as magnitude and another with a higher one.
    \item OLDA produces 2 events with really high magnitude, the highest of all methods, then 2 more with medium magnitude comparable to the one of MABED, and one with the lowest magnitude of all 3 methods.
    \item Peaky Topics produces 1 with really high magnitude, lower than OLDA but higher than MABED, and the rest of 4 with medium, medium-high magnitude where the lowest is higher than MABED's lowest.
\end{itemize}

For all proposed methods, the resulting topics are not easily understandable, as this is an issue addressed in the literature. 
In our experiments, all 3 methods use the whole window to generate topics.
To mitigate this shortcoming, the proposed EDSA-Ensamble architecture uses all 3 algorithms to detect results.

\subsection{Individual Sentiment Analysis}

As expected on a Sentiment Analysis task we notice that when using clean features, i.e., SFE text preprocessing, our models behave worse on all proposed models, deep and non-deep. 
All methods behave well on the standard dataset (above 75\% Accuracy and 70\% Precision) on all splits.
We notice that there is learning when moving from C1 to C2 and a small difference when moving from C2 to C3, more visible for the non-deep models as they are reaching their saturation point faster.
In terms of resources needed, LSTM consumes 10x more time than the standard non-deep models (all except SVM, where it is 3x more time), and in terms of needed memory and processing power way worse since the non-deep models were run perfectly on a CPU, and the deep ones needed GPU. 
Comparing the full training of the LSTM with the fine-tuning of the transformers is almost the same as comparing non-deep models with LSTM, but this time the increase in performance is not that high when it comes to C2 and C3, whereas LSTM performs best on C1.
There is almost 10\% difference between deep and non-deep models and deep ones where the latter reach almost 90\% Accuracy and Precision (89.2\% RoBERTa). 
Considering that the run-time is small even for transformers and the needed resources are not that big, for a server, the transformers are a viable model for this pipeline, but if we need to go with compact machines, LR will suffice.

When using EDSA-Ensemble, we observe an increase in accuracy for the individual sentiment analysis. 
This is expected as the misclassification of sentiment tweets is minimized by the ensemble model.

\subsection{Proposed Ensemble Solution}

In the majority of Sentiment Analysis tasks, there is little to no text pre-processing or cleaning done in order to maximize the effectiveness of the classifiers. 
Due to the nature of our solution, we have to prepare the dataset to run both sentiment analysis and event detection tasks so for that we did some pre-processing which might have lowered a bit the performance of the models on the separate task of Sentiment Analysis but have increased the performance on the entire EDSA-Ensamble solution.
We are benchmarking a variety of embeddings and ways of obtaining tokens in order to maximize the effectiveness of the proposed solution and we can notice that, when taken individually, Deep Learning models outperform classical Machine Learning
models, but not by far.
From our experiments, we observe that the cost (memory, processing power, training time) is not proportional to the results when training each individual model.
Our EDSA-Ensemble solution is mode resource-effective as it manages to increase the accuracy while its runtime is equal to the runtime of the model that takes the most.

On the task of Sentiment Analysis, we notice that LSTM and Transformers obtain comparable results but, on the end-to-end task  of detecting the sentiment of events, the transformers take the lead.
In terms of resources, the LSTM module is more effective and the performance is due to the fact that it is specifically trained for this task whereas the transformers are just fine-tuned for this task.

For the event detection task, we notice the EDSA-Ensemble model obtains good but different topics.
For a better understanding of events, it is better to extract events using multiple methods as we do with EDSA-Ensemble. 
In terms of general performance, the Perceptron models that employ Transformers for text preprocessing manage to obtain the best results, although they are outperformed by the proposed EDSA-Ensemble model.

\section{Conclusions}\label{Conclusions}

In this paper, we propose the EDSA-Ensemble model, a new approach that combines Event Detection and Sentiment Analysis for extracting the sentiments of events that appear in Social Networks.
By combining Network Analysis with Machine Learning, our solution brings together two otherwise isolated communities, i.e., Network Analysis and Natural Language Processing.

Event Detection algorithms manage to extract different bursty topics.
The experiment results show that OLDA increases topic readability and coherence, while MABED detects a wider range of topics.
For time-constrained and resource-limited use cases, Peaky Topics can also be considered, as the results are good for short intervals and comparable to the other two methods.
For a more accurate result, it is better to combine the results of these methods as we do with the EDSA-Ensamble model.

The evaluation of Sentiment Analysis algorithms provides a better inside into the construction of the model.
Although, with the increase of the dataset size the gap between the evaluation scores obtained by algorithms increases, this increase does not prove to be a real gain when taking into account the run-time.
Also depending on the chosen embedding, the models scale differently and might even not behave as desired.
Using an ensemble model that employs multiple models and text preprocessing and encoding techniques, the results of sentiment detection improve.

The proposed EDSA-Ensemble model for extracting the sentiments of the social events shows an improvement in performance over individual models. 
When taken individually, the overall best results are obtained when combining MABED with LSTM.
Although LSTM manages to improve accuracy significantly for Sentiment Analysis, this is done at a high cost of resources and time.
Furthermore, the experiments prove that the EDSA-Ensemble model manages to correctly determine the overall sentiment for an event. 

For future work, we aim to enhance EDSA-Ensamble with two Sentiment Analysis models based on Convolutional Neural Networks (CNN) and Generative Adversarial Networks (GAN), as well as, to extend the text preprocessing module with more word and transformer embedding methods, e.g., FastText~\cite{Bojanowski2017}, GloVe~\cite{Pennington2014}, MOE~\cite{Piktus2019}, BART~\cite{Lewis2020}, etc.
With EDSA-Ensemble, we also intend to identify communities that spread hate speech~\cite{Petrescu2021} and misinformation~\cite{Ilie2021,Truica2022,Truica2022a} and to mitigate this against this harmful content.

\bibliographystyle{plainnat}  
\bibliography{bibliography}

\end{document}